\documentclass[sigplan,twocolumn]{acmart}
\acmSubmissionID{999}

\usepackage[english]{babel}

\usepackage{blindtext}

\setcopyright{none}
\acmDOI{}
\acmISBN{}

\usepackage{tikz}
\usepackage{amsmath}

\usepackage{filecontents}

\usepackage{tikz}
\usepackage{xcolor}
\usepackage{xspace}
\usepackage{tikz}
\usepackage{pifont}
\usepackage[english]{babel}
\usepackage{blindtext}
\usepackage{lipsum}

\usepackage[ruled,vlined,linesnumbered]{algorithm2e}
\usepackage{booktabs}

\usepackage{filecontents}
\usepackage{xspace}
\usepackage{subfig}
\usepackage{xcolor}
\usepackage{hhline}
\usepackage{multirow}
\usepackage{soul}
\newtheorem{definition}{Definition}
\usepackage{array}
\usepackage{comment}
\usepackage{listings}
\usepackage{dsfont}
\usepackage{algorithmic}
\usepackage{xurl}
\usepackage{setspace}
\usepackage[flushleft]{threeparttable}
\usepackage{caption}
\usepackage{ltablex}

\usepackage{ifluatex}

\usepackage{outlines}

\usepackage{multirow}

\usepackage{hyperref}
\usepackage{hyperxmp}

\newtheorem{insight}{Insight}
\newtheorem*{goal}{Goal}

\begin{document}

\newcommand*{\affmark}[1][*]{\textsuperscript{#1}}
\newcommand*{\affaddr}[1]{#1}

\acmYear{2025}\copyrightyear{2025}
\setcopyright{acmlicensed}
\acmConference[EuroSys '25]{Twentieth European Conference on Computer Systems}{March 30--April 3, 2025}{Rotterdam, Netherlands}
\acmBooktitle{Twentieth European Conference on Computer Systems (EuroSys '25), March 30--April 3, 2025, Rotterdam, Netherlands}
\acmDOI{10.1145/3689031.3696098}
\acmISBN{979-8-4007-1196-1/25/03}

\title{{\name: Fast Large Language Model Serving for RAG with Cached Knowledge Fusion}}



\author{Jiayi Yao}
\affiliation{%
  \institution{University of Chicago/CUHK Shenzhen}}

\author{Hanchen Li}
\affiliation{%
  \institution{University of Chicago}}

\author{Yuhan Liu}
\affiliation{%
  \institution{University of Chicago}}

\author{Siddhant Ray}
\affiliation{%
  \institution{University of Chicago}}

\author{Yihua Cheng}
\affiliation{%
  \institution{University of Chicago}}

\author{Qizheng Zhang}
\affiliation{%
  \institution{Stanford University}}

\author{Kuntai Du}
\affiliation{%
  \institution{University of Chicago}}

\author{Shan Lu}
\affiliation{%
  \institution{Microsoft Research / University of Chicago}}  

  \author{Junchen Jiang}
\affiliation{%
  \institution{University of Chicago}}

\keywords{Large Language Models, KV Cache, Retrieval-Augmented-Generation}

\ccsdesc[500]{Computing methodologies~Natural language processing}
\ccsdesc[500]{Networks~Cloud computing}
\ccsdesc[300]{Information systems~Data management systems}

\newcommand{\edit}[1]{{\color{black} #1}}
\newcommand{\shan}[1]{{\color{red}(Shan: #1)}}
\newcommand{\shanedit}[1]{{\color{red} #1}} 
\newcommand{\hank}[1]{{\color{blue}(Hank: #1)}}
\newcommand{\mm}[1]{{\color{violet}(Michael:#1)}}
\newcommand{\jc}[1]{{\footnotesize\color{orange}{(JC: #1)}}}
\newcommand{\jcedit}[1]{{\color{orange} #1}} 
\newcommand{\yh}[1]{{\footnotesize\color{deepgreen}{(Yuhan: #1)}}}
\newcommand{\jiayi}[1]{{\color{brown}{(Jiayi: #1)}}}
\newcommand{\todo}[1]{{\color{red}{(TODO: #1)}}}
\newcommand{\hcedit}[1]{{\color{black} #1}}
\definecolor{darkkhaki}{rgb}{0.74, 0.72, 0.42}
\newcommand{\hc}[1]{{\color{darkkhaki}{(LHC: #1)}}}
\definecolor{hhcolor}{RGB}{240, 35, 240}
\newcommand{\hh}[1]{{\color{hhcolor}{(Yihua: #1)}}}
\newcommand{\sr}[1]{{\color{cyan!70!blue}{(Siddhant: #1)}}}
\newcommand{\qz}[1]{{\color{purple}{(Qizheng: #1)}}}

\newcommand{\KV}{\ensuremath{KV}\xspace}
\newcommand{\Attention}{\ensuremath{A}\xspace}
\newcommand{\ForwardAttention}{\ensuremath{FA}\xspace}
\newcommand{\LayerIndex}{\ensuremath{i}\xspace}
\newcommand{\TokenIndex}{\ensuremath{j}\xspace}
\newcommand{\UserTokenIndex}{\ensuremath{j'}\xspace}
\newcommand{\ChunkIndex}{\ensuremath{n}\xspace}
\newcommand{\ChunkNum}{\ensuremath{N}\xspace}
\newcommand{\Func}{\ensuremath{F}\xspace}
\newcommand{\Pre}{\textrm{pre}\xspace}
\newcommand{\New}{\textrm{new}\xspace}
\newcommand{\Full}{\textrm{full}\xspace}
\newcommand{\CA}{\ensuremath{C}\xspace}

\newcommand{\KVD}{\ensuremath{\Delta_{\textrm{kv}}}\xspace}
\newcommand{\CAD}{\ensuremath{\Delta_{\textrm{attn}}}\xspace}

\newcommand{\ACA}{{ACA}\xspace}
\newcommand{\HCA}{{HKVD}\xspace}
\newcommand{\name}{\textsc{CacheBlend}\xspace}
\newcommand{\promptcache}{PromptCache\xspace}
\newcommand{\HCF}{HCF tokens\xspace}

\newcommand{\vspacesize}{0.2cm}

\newcommand{\fillme}{{\bf XXX}\xspace}

\newcommand*\circled[1]{\tikz[baseline=(char.base)]{
            \node[shape=circle,fill,inner sep=2pt] (char) {\textcolor{white}{\footnotesize{#1}}};}}

\newcounter{packednmbr}
\newenvironment{packedenumerate}{\begin{list}{\thepackednmbr.}{\usecounter{packednmbr}\setlength{\itemsep}{0.5pt}\addtolength{\labelwidth}{-4pt}\setlength{\leftmargin}{2ex}\setlength{\listparindent}{\parindent}\setlength{\parsep}{1pt}\setlength{\topsep}{0pt}}}{\end{list}}
\newenvironment{packeditemize}{\begin{list}{$\bullet$}{\setlength{\itemsep}{0.5pt}\addtolength{\labelwidth}{-4pt}\setlength{\leftmargin}{2ex}\setlength{\listparindent}{\parindent}\setlength{\parsep}{1pt}\setlength{\topsep}{2pt}}}{\end{list}}
\newenvironment{packedpackeditemize}{\begin{list}{$\bullet$}{\setlength{\itemsep}{0.5pt}\addtolength{\labelwidth}{-4pt}\setlength{\leftmargin}{\labelwidth}\setlength{\listparindent}{\parindent}\setlength{\parsep}{1pt}\setlength{\topsep}{0pt}}}{\end{list}}
\newenvironment{packedtrivlist}{\begin{list}{\setlength{\itemsep}{0.2pt}\addtolength{\labelwidth}{-4pt}\setlength{\leftmargin}{\labelwidth}\setlength{\listparindent}{\parindent}\setlength{\parsep}{1pt}\setlength{\topsep}{0pt}}}{\end{list}}
\let\enumerate\packedenumerate
\let\endenumerate\endpackedenumerate
\let\itemize\packeditemize
\let\enditemize\endpackeditemize

\newcommand{\tightcaption}[1]{\vspace{-0.2cm}\caption{{\normalfont{\textit{{#1}}}}}\vspace{-0.2cm}}
\newcommand{\tightsection}[1]{\vspace{-0.3cm}\section{#1}\vspace{-0.2cm}}
\newcommand{\tightsectionstar}[1]{\vspace{-0.17cm}\section*{#1}\vspace{-0.08cm}}
\newcommand{\tightsubsection}[1]{\vspace{-0.25cm}\subsection{#1}\vspace{-0.1cm}}
\newcommand{\tightsubsubsection}[1]{\vspace{-0.01in}\subsubsection{#1}\vspace{-0.01cm}}

\newcommand{\eg}{{\it e.g.,}\xspace}
\newcommand{\ie}{{\it i.e.,}\xspace}
\newcommand{\etal}{{\it et.~al}\xspace}
\newcommand{\bigO}{\mathrm{O}}
\newcommand{\twlog}{w.l.o.g.\xspac}

\newcommand{\myparashort}[1]{\vspace{0.05cm}\noindent{\bf {#1}}~}
\newcommand{\mypara}[1]{\vspace{0.05cm}\noindent{\bf {#1}:}~}
\newcommand{\myparatight}[1]{\vspace{0.02cm}\noindent{\bf {#1}:}~}
\newcommand{\myparaq}[1]{\smallskip\noindent{\bf {#1}?}~}
\newcommand{\myparaittight}[1]{\smallskip\noindent{\emph {#1}:}~}
\newcommand{\question}[1]{\smallskip\noindent{\emph{Q:~#1}}\smallskip}
\newcommand{\myparaqtight}[1]{\smallskip\noindent{\bf {#1}}~}

\newcommand{\cmark}{\ding{51}}%
\newcommand{\xmark}{\ding{55}}%



\definecolor{backcolour}{rgb}{0.96,0.96,0.96}
\definecolor{codegray}{rgb}{0.5,0.5,0.5}
\definecolor{deepblue}{rgb}{0,0,0.6}
\definecolor{deepred}{rgb}{0.6,0,0}
\definecolor{deepgreen}{rgb}{0,0.5,0}
\lstdefinestyle{mystyle}{
    backgroundcolor=\color{backcolour},   
    commentstyle=\color{codegreen},
    morekeywords={self, True},
    keywordstyle=\color{deepblue},
    numberstyle=\tiny\color{codegray},
    emph={MyClass,__init__,EncodingType,Image},
    emphstyle=\color{deepred},
    stringstyle=\color{deepgreen},
    basicstyle=\ttfamily\footnotesize,
    breakatwhitespace=false,         
    breaklines=true,                 
    captionpos=b,                    
    keepspaces=true,                 
    numbers=left,                    
    numbersep=5pt,                  
    showspaces=false,                
    showstringspaces=false,
    showtabs=false,                  
    tabsize=1
}


\begin{abstract}

Large language models (LLMs) often incorporate multiple text chunks in their inputs to provide the necessary contexts. 
To speed up the prefill of the long LLM inputs, one can {\em pre}-compute the KV cache of a text and {\em re}-use the KV cache when the context is reused as the prefix of another LLM input.
However, the reused text chunks are {\em not} always the input prefix, which makes precomputed KV caches not directly usable since they ignore the text's {\em cross-attention} with the preceding texts.
Thus, the benefits of reusing KV caches remain largely unrealized.

This paper tackles just one challenge: when an LLM input contains {\em multiple} text chunks, {\em how to quickly combine their precomputed KV caches} in order to achieve the same generation quality as the expensive full prefill (\ie without reusing KV cache)?
This challenge naturally arises in retrieval-augmented generation (RAG) where the input is supplemented with multiple retrieved texts as the context.
We present \name, a scheme that reuses the pre-computed KV caches, regardless prefix or not, and {\em selectively recomputes the KV values of a small subset of tokens} to partially update each reused KV cache.
In the meantime, the small extra delay for recomputing some tokens can be pipelined with the retrieval of KV caches within the same job, allowing \name to store KV caches in slower devices with more storage capacity while retrieving them
{\em without} increasing the inference delay. 
By comparing \name with the state-of-the-art KV cache reusing schemes on three open-source LLMs of various sizes and four popular benchmark datasets of different tasks, we show that \name reduces time-to-first-token (TTFT) by 2.2--3.3$\times$ and increases the inference throughput by 2.8-5$\times$ from full KV recompute without compromising generation quality. The code is available at \href{https://github.com/LMCache/LMCache}{\textcolor{blue}{\textbf{https://github.com/LMCache/LMCache}}}.
\end{abstract}

\maketitle
\pagestyle{plain} 


\section{Introduction}

For their remarkable capabilities, large language models (LLMs) are widely used in personal assistance, AI healthcare, and question answering~\cite{llm-app-1,llm-app-2,llm-app-3,llm-app-4}. 
To ensure high-quality and consistent responses, applications often supplement the user query with additional texts to provide the necessary {\em context} of domain knowledge or user-specific information.
A typical example is Retrieval-Augmented Generation (RAG) where
a user query will be prepended by multiple {\em text chunks} retrieved from a database to form the LLM input. 

\begin{figure}
\centering
     \includegraphics[width=.96\columnwidth]{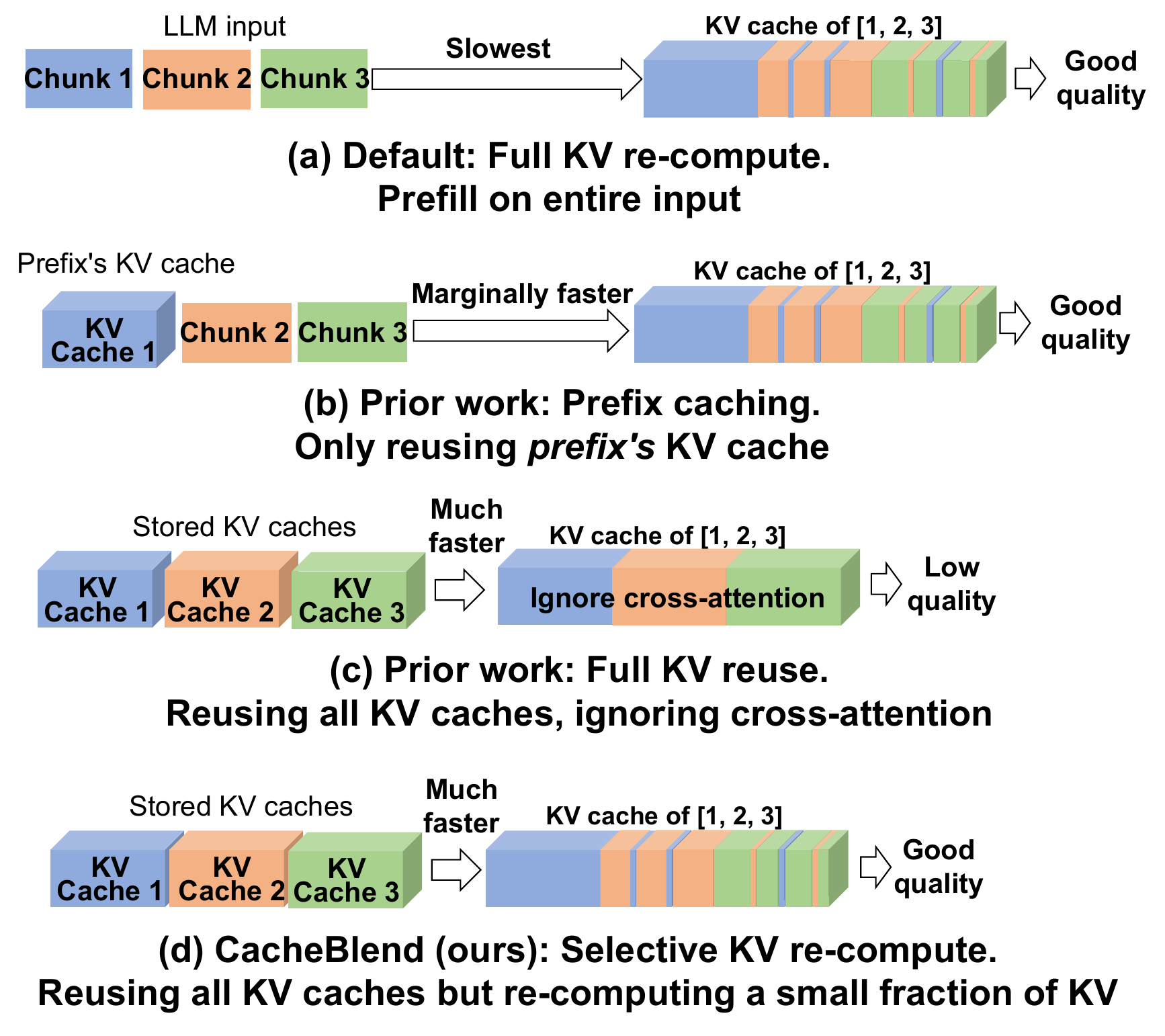}
    \tightcaption{Contrasting full KV recompute, prefix caching, full KV reuse, and \name's selective KV recompute.}
    \label{fig:intro:contrast}
\end{figure}

These context text chunks, however, significantly slow down LLM inference. 
This is because, before generating any token, an LLM first uses {\em prefill} to go through the entire LLM input to produce the {\em KV cache}---concatenation of tensors associated with each input token that embeds the token's ``attention'' with its preceding tokens.
Thus, the prefill delay determines the time to first token (TTFT).
We refer to it as {\em full KV recompute} (Figure~\ref{fig:intro:contrast}(a)).
Despite many optimizations, 
the delay and computation of prefill grow super-linearly with the input length, and can easily slow down the service, especially on long LLM inputs (\eg in RAG)~\cite{zhong2024distserve,agrawal2023sarathi,wu2024loongserve}.

So, how do we speed up the prefill of LLM inputs?
Recent optimizations embrace the fact that same context texts are often reused by different LLM inputs. 
They then {\em pre}-compute the KV caches of these texts {\em once} and {\em re}-use the stored KV caches to avoid repeated prefill on these reused texts. 

\mypara{Limitations of existing methods}
There are currently two approaches to KV cache reusing, but they both have limitations.

First, {\em prefix caching} only stores and reuses the KV cache of the prefix of the LLM input~\cite{cachegen,sglang,kwon2023efficient,jin2024ragcache,liu2024optimizingsql} (Figure~\ref{fig:intro:contrast}(b)).
Because the prefix's KV cache is independent of the succeeding texts, prefix caching does not hurt generation quality.
However, many applications, such as RAG, include {\em multiple} text chunks, rather than one, in the LLM input to provide all necessary contexts to ensure good response quality. 
Thus, only the first text chunk is the prefix, and other reused texts' KV caches are not reused.
As a result, the speed of prefix caching will be almost as slow as full KV recompute when input consists of many reused text chunks.


Second, {\em full KV reuse} aims to address this shortcoming (Figure~\ref{fig:intro:contrast}(c)).
When a reused text is not at the input prefix, it still reuses the KV cache by adjusting its positional embedding so that the LLM generation will produce meaningful output~\cite{gim2023prompt}. 
However, this method ignores the important {\em cross-attention}---the attention between tokens in one chunk with tokens in its preceding chunks. 
The cross-attention information cannot be pre-computed as the preceding chunks are not known in advance.
Yet, cross-attention can be vital to answer queries (\eg about geopolitics) that naturally require understanding information from multiple chunks {\em jointly} (\eg chunks about geography and chunks about politics). 
\S\ref{subsec:modular} offers concrete examples to illustrate when prefix caching and modular caching are insufficient.

\mypara{Our approach}
This paper tackles only one challenge: when an LLM input includes multiple text chunks, how to {\em quickly} combine their individually pre-computed KV caches, in order to achieve the same generation {\em quality} as the expensive full prefill?
In other words, we seek to have both the speed of {\em full KV reuse} and the generation quality of {\em full KV recompute}. 

We present \name, a system that fuses multiple pre-computed KV caches, regardless of prefix or not, by {\em selectively recomputing} the KV cache of a small fraction of tokens, based on the preceding texts in the specific LLM input.
We refer to it as {\bf selective KV recompute} (Figure~\ref{fig:intro:contrast}(d)). 
At a high level, selective KV recompute performs prefill on the input text in a traditional layer-by-layer fashion; however, in each layer, it updates the KV of only a small fraction of tokens while reusing the KV of other tokens.

Comparing with full KV recompute,
an update fraction of less than 15\% can typically
generate same-quality responses based on our experience.
The deeper reason why it suffices to only updating a small fraction of KV is due to the sparsity of attention matrices (see \S\ref{subsec:token-selection}).

Comparing with full KV reuse, \name achieves
higher generation quality with a small amount
of extra KV update. 
Moreover, this small extra computation does not increase the inference latency, 
because \name parallelizes partial KV update on one layer with the fetching of the KV cache on the next layer into GPU memory.
Such pipelining enables \name to store KV caches in slower non-volatile devices (\eg disk) {\em without} incurring extra delay, allowing more KV caches to be stored and reused. 

To put our contribution in context, \name enables the reusing of KV caches of multiple text chunks in one LLM input, without compromising generation quality.
This is complementary to the recent work that reduces KV cache storage sizes~\cite{h2o,liu2024scissorhands,cachegen, kvquant, kivi, gear} and optimizes the access patterns of KV cache~\cite{sglang,jin2024ragcache}. 

We implemented \name on top of vLLM and compared \name with state-of-the-art KV cache reusing schemes on three open-source LLMs of various sizes and three popular benchmark datasets of two LLM tasks (RAG and QA).
We show that compared to prefix caching, \name reduces time-to-first-token (TTFT) by 2.2--3.3$\times$ and increases the inference throughput by 2.8--5$\times$, without compromising generation quality or incurring more storage cost.
Compared to full KV reuse, 
\name achieves almost the same TTFT but 0.1-0.2 higher absolute F1-scores on QA tasks and 0.03-0.25 higher absolute Rouge-L scores on summarization.

\vspace{-4pt}
\section{Background}
\label{sec:background}



Most LLM services today use transformers~\cite{vaswani2023attention,brown2020language, palm}.
After receiving the input tokens, the LLM first uses the prefill phase (explained shortly) to transform the tokens into key (K) and value (V) vectors, \ie~{\em KV cache}.
After prefill, the LLM then
iteratively decodes (generates)
the next token with the current KV cache and appends the new K and V vectors of the new tokens to the KV cache for the next iteration. 

The prefill phase computes the KV cache layer by layer.
The input tokens embeddings on each layer are first transformed into query (Q), key (K), and value (V) vectors, of which the K and V vectors form one layer of the KV cache.
The LLM then multiplies Q and K vectors to obtain the {\em attention matrix}---the attention between each token and its preceding tokens---and does another dot product between the (normalized and masked) attention matrix with the V vector. 
The resulting vector will go through multiple neural layers to obtain the tokens' embeddings on the next layer. 

When the KV cache of a prefix is available, the prefill phase will only need to compute the {\em forward attention} matrix (between the suffix tokens and the prefix tokens) on each layer which directly affects the generated token. 


The prefill phase can be slow, especially on long inputs. 
For instance, on an input of four thousand tokens (a typical context length in RAG~\cite{jin2024ragcache}), running prefill can take three (or six) seconds for Llama-34B (or Llama-70B) on one A40 GPU.
This causes a substantial delay that users have to wait before seeing the first word generated.
Recent work also demonstrates that prefill can be a throughput bottleneck, by showing that getting rid of the prefill phase can double the throughput of an LLM inference system~\cite{zhong2024distserve}.

\vspace{-10pt}
\section{Motivation}
\subsection{Opportunities of reusing KV caches}
 \label{subsec:redundancy}

Recent systems try to alleviate the prefill overhead by leveraging the observation that in many LLM use cases, the same texts are used repeatedly in different LLM inputs. 
This allows reusing KV caches of these reused texts (explained shortly). 

Text reusing is particularly prevalent when same texts are included in the LLM input 
to provide necessary contexts to ensure high and consistent response quality. 
To make it more concrete, let's consider two scenarios.
\begin{packeditemize}
\item In a company that uses LLM to manage internal records, two queries can be 
{\em ``who in the IT department proposed using RAG to enhance the customer service X during the last all-hands meeting?''} and {\em ``who from the IT department were graduates from college Y?''}
While seemingly different, both queries involve {\em the list of employees in the IT department} as a necessary context to generate correct answers. 
\item Similarly, in an LLM-based application that summarizes Arxiv papers, two queries can be 
{\em ``what are the trending RAG techniques on Arxiv?''} and {\em ``what datasets are used recently to benchmark RAG-related papers on Arxiv?''} 
They both need the {\em recent Arxiv papers about RAG} as the necessary context to generate correct results.
\end{packeditemize}

Since the reused contexts typically contain more information than the user queries, the prefill on the ``context'' part of the input accounts for the bulk of prefill overhead~\cite{jin2024ragcache, gao2023retrievalaugmented}.
Thus, it would be ideal to store and reuse the KV caches of reused texts, in order to avoid the prefill overhead when these texts are used again in different LLM inputs.




\subsection{Why is prefix caching insufficient?}
\label{subsec:prefixcache}

Indeed, several recent systems are developed to reduce prefill delay by reusing KV caches. 
For example, in prefix caching, the KV cache of a reusable text chunk is precomputed once, and if the text chunk is at the {\em prefix} of an LLM input, then the precomputed KV cache can be reused to avoid prefill on the prefix. 
The advantage of prefix caching is that the KV cache of a prefix is not affected by the succeeding text, so the generation result will be identical to full KV recompute (without the KV cache). 
Several systems have followed this approach, \eg vLLM~\cite{kwon2023efficient}, SGLang~\cite{sglang}, and RAGCache~\cite{jin2024ragcache}.

The disadvantage of prefix caching is also clear.
To answer one query, applications, such as RAG, often prepend {\bf \em multiple} text chunks in the LLM input to provide different contexts necessary for answering the query.\footnote{Prepending all contexts in an LLM input is a popular way of augmenting user queries (\eg ``stuff'' mode in Langchain and LlamaIndex). 
By default, this paper uses this mode.
Other RAG methods like ``MapReduce'' or ``Rerank'' do not fit in this category.
They first process each context text chunk separately before combining the results from each context. Since each chunk is always the prefix when processed separately, prefix caching works well. 
However, ``MapReduce'' is slow since it needs to summarize every chunk before generating the answer from summaries.  
``Rerank'' suffers from low generation quality if multiple chunks contain relevant information as it processes every chunk individually. 
We also empirically evaluate these methods and compare them with \name in \S\ref{sec:eval}.
}
As a result, {\bf \em except the first chunk, all other chunks' KV caches are not reused since they are not the prefix of the LLM input}.

Let us think about the queries from \S\ref{subsec:redundancy}. 
To answer ``{\em who in the IT department proposed using RAG to enhance the customer service X during the last all-hands meeting?}'', we need contexts from {\em multiple} sources, including IT department's employees, information about service X, and meeting notes from the all-hands meeting. 
Similarly, in the Arxiv-summarization app, answering the example queries will require the LLM to read {\em several} recent RAG-related Arxiv papers as the contexts. 
Being on different topics, these contexts are unlikely to appear together in one text chunk. They are separate text chunks used together only when answering a particular query. 


\begin{figure}
\centering
     {\includegraphics[width=.99\columnwidth]{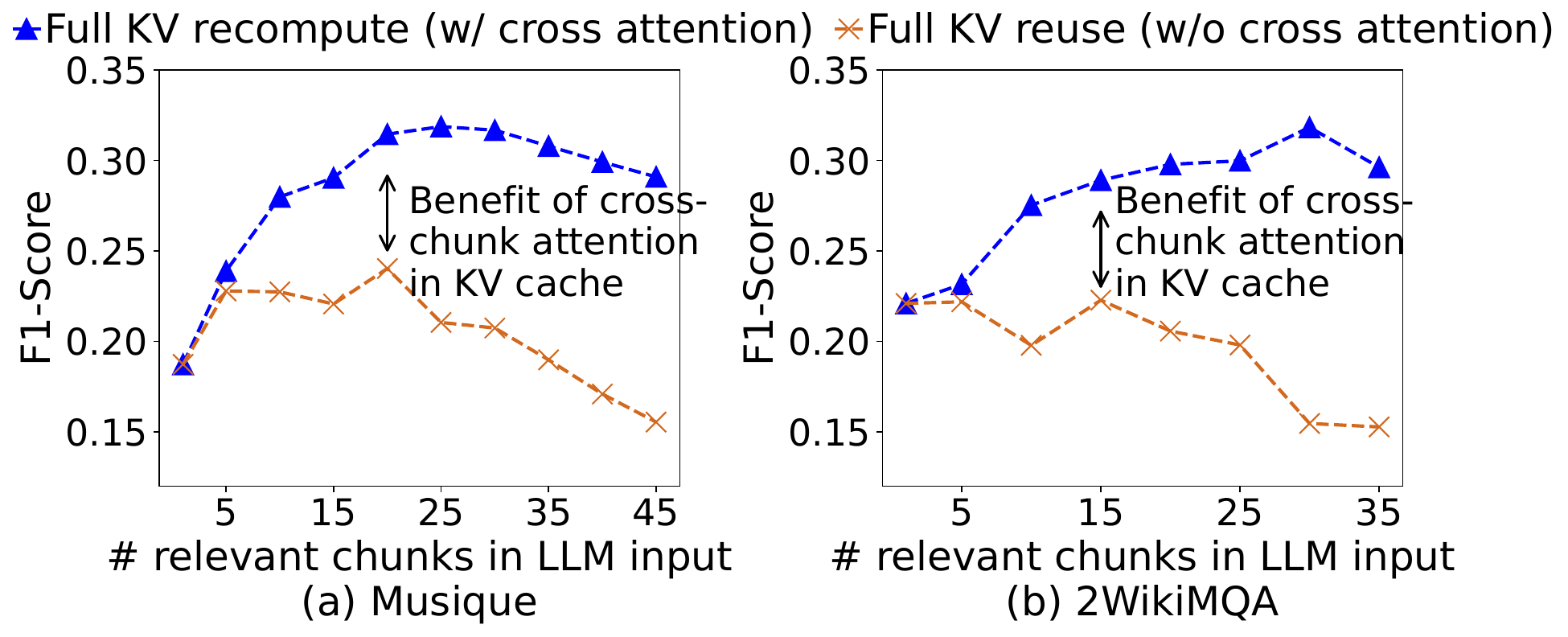}}\label{fig:rag_acc_1}
    \tightcaption{Generation quality improves as more text chunks are retrieved. 
    }
    
    \label{fig:more-chunks}
\end{figure}

To empirically show the needs for including multiple text chunks in LLM inputs, we use two popular multi-hop QA datasets, Musique and 2WikiMQA. These datasets consist of queries and multiple associated context texts needed to answer the queries. 
Following the common practice of RAG, we first create a vector database by splitting the contexts into chunks of 128 tokens (a popular number~\cite{jan_chunksize}) using the text chunking mechanism from Langchain~\cite{langchain}. 
For each query, we embed the query using SentenceTransformers~\cite{reimers-2019-sentence-bert}, and fetch top-k relevant chunks from the database, based on the least L2 distance between the embeddings of the query and the chunk respectively.
Figure~\ref{fig:more-chunks} shows the generation quality, measured using a standard \emph{F1-score} metric, with an increasing number of selected text chunks.
We can see that the quality improves significantly as more text chunks are retrieved to supplement the LLM input, though including too many chunks hurts quality due to the well-known lost-in-the-middle issue~\cite{liu2023lost,xu2023retrievallost}.

In short, {\em prefix caching can only save the prefill of the first text chunk}, so the saving will be marginal when the LLM input includes more text chunks, even if they are reused.

\subsection{Why is full KV reuse insufficient?}
\label{subsec:modular}

Full KV reuse is proposed to address this very problem. 
This approach is recently pioneered by PromptCache~\cite{gim2023prompt}.
It concatenates independently precomputed KV caches of recurring text chunks with the help of {\em buffers} to maintain the positional accuracy of each text chunk.
For instance, to concatenate the KV caches of chunks \(C_{1}\) and \(C_{2}\), PromptCache first needs to precompute the KV cache of \(C_{2}\) by running prefill on a hypothetical input that prepends \(C_{2}\) with a dummy prefix 
of length greater or equal to \(C_{1}\).
This way, even if \(C_{2}\) is not the prefix, we still correctly preserve the positional information of \(C_{2}\)'s KV cache, though each chunk's KV cache will have to be precomputed multiple times.


However, even with the positional information preserved, a more fundamental problem is that the KV cache of non-prefix text chunk (\eg \(C_{2}\)) {\bf \em ignores the cross-attention} between the chunk and the preceding text (\eg \(C_{1}\)).
This is because the preceding text is not known when precomputing the KV cache. 

\begin{figure}
\centering
     \includegraphics[width=.99\columnwidth]{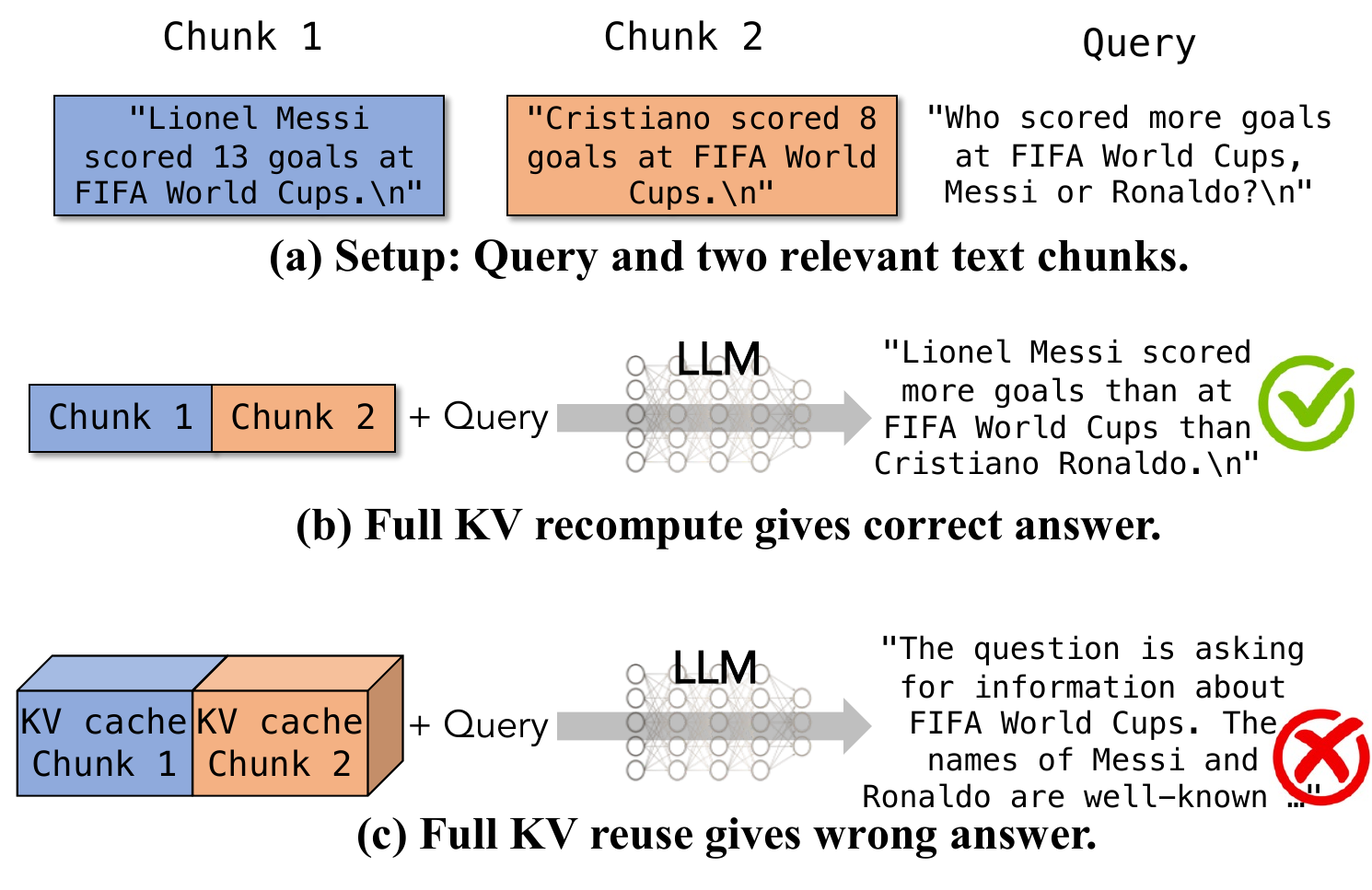}
    \tightcaption{An illustrative example of an LLM input with two text chunks prepended to a query. 
    Full KV recompute (b), without reusing KV cache, is slow but gives the correct answer. 
    Full KV reuse (c), however, gives the wrong answer as it neglects cross-attention between the chunks (Figure~\ref{fig:attn_comparison}).} 
    \label{fig:text_comparison}
\end{figure}

Ignoring cross-attention can lead to a wrong response. 
Figure~\ref{fig:text_comparison} shows an illustrative example, where a user query \textit{"How many goals did Messi score more than Cristiano Ronaldo at FIFA World Cups?"} is prepended by the two text chunks of the players' career statistics. 
With full prefill or prefix caching, the result is clear and correct. 
With full KV reuse the KV caches of the two text chunks are precomputed, with each chunk having the right positional embedding, and then concatenated to form the KV cache.
However, if the LLM uses this KV cache to generate the answer, it will start to ramble and not produce the right answer.

\begin{figure}
\centering
     \includegraphics[width=.92\columnwidth]{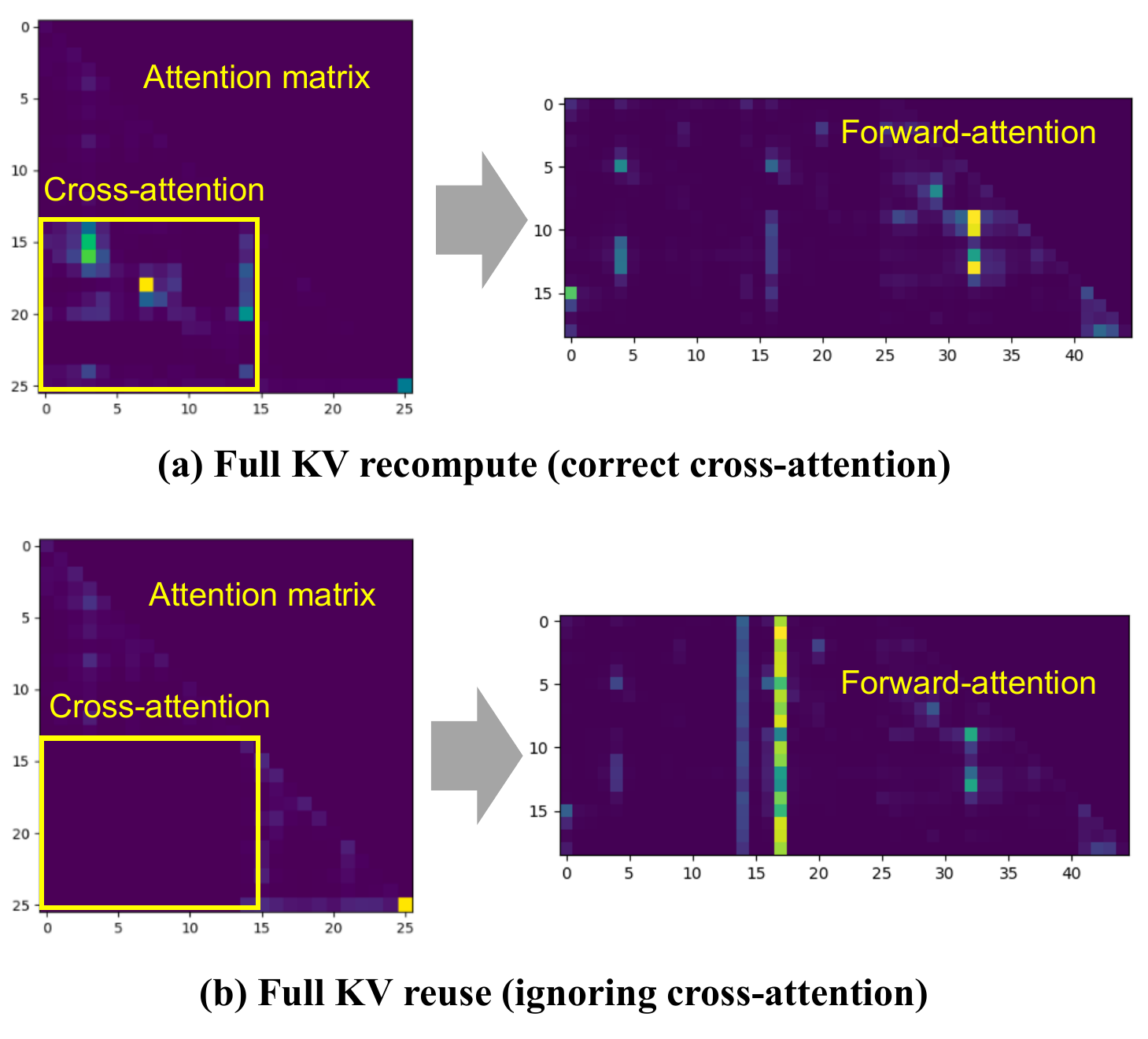}
     \vspace{-3pt}
    \tightcaption{Contrasting the attention matrices of (a) full KV recompute and (b) full KV reuse. The yellow boxes highlight the cross-attention. The right-hand side plots show the resulting forward attention matrices whose discrepancies are a result of the different cross-attention between the two methods.} 
    \label{fig:attn_comparison}
    \vspace{-5pt}
\end{figure}

To understand why, we take a closer look at the {\bf attention matrix} (explained in \S\ref{sec:background}), particularly the cross-attention between the two text chunks talking about the players' statistics. 
Figure~\ref{fig:attn_comparison} visualizes the attention matrix resulting from the KV cache of the original (full) prefill and the KV cache of full KV reuse.
Since full KV reuse precomputes each chunk separately, the cross attention between two chunks is completely missed (never computed) when the KV caches are precomputed. 
In this example, the first chunk contains Messi's goal count and the second chunk contains Ronaldo's. 
The LLM is queried to compare the goal counts between Messi and Ronaldo. 
Neglecting the interaction (cross-attention) between two chunks would lead to a flawed answer.

In fairness, it should be noted that full KV reuse does work when the cross-attention between chunks is low. 
This can commonly occur with prompt templates which are the main target application of PromptCache~\cite{gim2023prompt}.

The absence of cross-attention in full KV reuse causes significant discrepancies in the {\bf forward attention} matrix (explained in \S\ref{sec:background}), which contains the attention between context tokens and the last few tokens, and directly affects the generated tokens.

To show the prevalence of cross-attention in multi-chunk LLM inputs, Figure~\ref{fig:more-chunks} contrasts the response quality (in F1 score) between full KV recompute (with cross-attention) and full KV reuse (without cross-attention). 
We can see that as the number of relevant chunks increases, the disparity between full prefill and modular caching becomes more pronounced. This is because, with a larger number of chunks, the amount of cross-referencing and interdependency between different parts of the input (cross-attention) increases.

\vspace{-5pt}
\section{Fast KV Cache Fusing}
\label{sec:merge}

Given that {\em full KV recompute} (\ie full prefill or prefix caching) can be too slow while {\em full KV reuse} has low quality, a natural question then is how to have both the speed of full KV reuse {\em and} the quality of full KV recompute. 
Our goal, therefore, is the following:

\begin{goal}
When an LLM input includes multiple re-used text chunks, how to {\bf quickly} update the pre-computed KV cache, such that the forward attention matrix (and subsequently the output text) has {\bf minimum difference} with the one produced by full KV recompute. 
\end{goal}

To achieve our goal, we present \name, which {\em recomputes the KV of a selective subset of tokens} on each layer while reusing other tokens' KV.\footnote{For simplicity, we use the terms KV and KV cache interchangeably.}
This section explains \name in three parts.
We begin with the notations (\S\ref{subsec:terminology}), and then describe {\em how to recompute} the KV of only a small subset of tokens (\S\ref{subsec:partial-recompute}), and finally
explain {\em how to select} the tokens on each layer whose KV will be recomputed (\S\ref{subsec:token-selection}).

\begin{table}[]
\small
\begin{tabular}{rl}
\hline
\textit{Notation} & \textit{Description} \\ \hline\hline
\begin{tabular}[c]{@{}r@{}}$\LayerIndex$ \\ $\TokenIndex$\end{tabular} & \begin{tabular}[c]{@{}l@{}}Layer index\\ Token index\end{tabular} \\ \hline
\begin{tabular}[c]{@{}r@{}}$\KV$\\ $\KV_{\LayerIndex}$\\ $\KV_{\LayerIndex}[\TokenIndex]$\\ $\KV^{\Full}$\\ $\KV^{\Pre}$\\ $\KV^{\New}$\end{tabular} & \begin{tabular}[c]{@{}l@{}}KV cache\\ KV on layer $\LayerIndex$\\ KV on layer $\LayerIndex$ at token $\TokenIndex$\\ Fully recomputed KV cache\\ Pre-computed KV cache\\ \name-updated KV cache\end{tabular} \\ \hline
\begin{tabular}[c]{@{}r@{}}$\Attention_{\LayerIndex}$\\ $\Attention_{\LayerIndex}^{\Full}$\\ $\Attention_{\LayerIndex}^{\Pre}$\\ $\Attention_{\LayerIndex}^{\New}$\end{tabular} & \begin{tabular}[c]{@{}l@{}}Forward attention matrix on layer $\LayerIndex$\\ Forward attention matrix of full KV recompute\\ Forward attention matrix of full KV reuse\\ Forward attention matrix with \name\end{tabular} \\ \hline
$\KVD(\KV_{\LayerIndex}, \KV^{\Full}_{\LayerIndex})[\TokenIndex]$ & KV deviation between $\KV_{\LayerIndex}[\TokenIndex]$ and $\KV_{\LayerIndex}^{\Full}[\TokenIndex]$ \\ \hline
$\CAD(\Attention_{\LayerIndex},\Attention_{\LayerIndex}^{\Full})$ & Attention deviation between $\Attention_{\LayerIndex}$ and $\Attention_{\LayerIndex}^{\Full}$ \\ \hline
\end{tabular}
\caption{Summary of terminology}
\label{tab:terminology}
\vspace{-15pt}
\end{table}

\subsection{Terminology}
\label{subsec:terminology}

Table~\ref{tab:terminology} summarizes the notations used in this section.
For a given list of $\ChunkNum$ text chunks, 
we use $\KV^{\Full}$ to denote the KV cache from full KV recompute, $\KV^{\Pre}$ to denote the pre-computed KV cache,  and $\KV^{\New}$ to denote the \name-updated KV cache. 
Here, each of these KV caches is a concatenation of KV caches associated with different text chunks.
Each layer $\LayerIndex$ of the KV cache, $\KV_{\LayerIndex}$, produces the forward attention matrix $\Attention_{\LayerIndex}$.


The difference between the full KV recompute (full prefill) and the full KV re-use is two-fold. 

\begin{packeditemize}
\item {\bf KV deviation:}
We define the {\em KV deviation} of a KV cache $\KV$ on layer $\LayerIndex$ of token $\TokenIndex$ as the absolute difference between $\KV_{\LayerIndex}[\TokenIndex]$ and $\KV_{\LayerIndex}^{\Full}[\TokenIndex]$, denoted as $\KVD(\KV_{\LayerIndex}, \KV^{\Full}_{\LayerIndex})[\TokenIndex]$. 
It measures how much different the given KV is on a particular token and layer compared to the full-prefilled KV cache.
We will later use the KV deviation to identify which tokens' KV has higher deviation and thus need to be updated. 

\item {\bf Attention deviation:}
Similarly, for the forward attention matrix $\Attention_{\LayerIndex}$ on layer $\LayerIndex$, we define the {\em attention deviation}, denoted as $\CAD(\Attention_{\LayerIndex}, \Attention_{\LayerIndex}^{\Full})$, to be the L-2 norm of its difference with $\Attention_{\LayerIndex}^{\Full}$.
Recall from \S\ref{subsec:modular} that full KV reuse suffers from deviation in the forward attention matrix (illustrated in Figure~\ref{fig:attn_comparison}) due to the absence of cross-attention.


\end{packeditemize}


Using these notations, our goal can be formulated as how to quickly update the precomputed KV cache $\KV^{\Pre}$ to the new KV cache $\KV^{\New}$, such that the attention deviation $\CAD(\Attention^{\New}_{\LayerIndex},\Attention^{\Full}_{\LayerIndex})$ on any layer $\LayerIndex$, is minimized.

\subsection{Selectively recomputing KV cache}
\label{subsec:partial-recompute}

\begin{figure}[t]
\centering
	\includegraphics[width=.99\columnwidth]{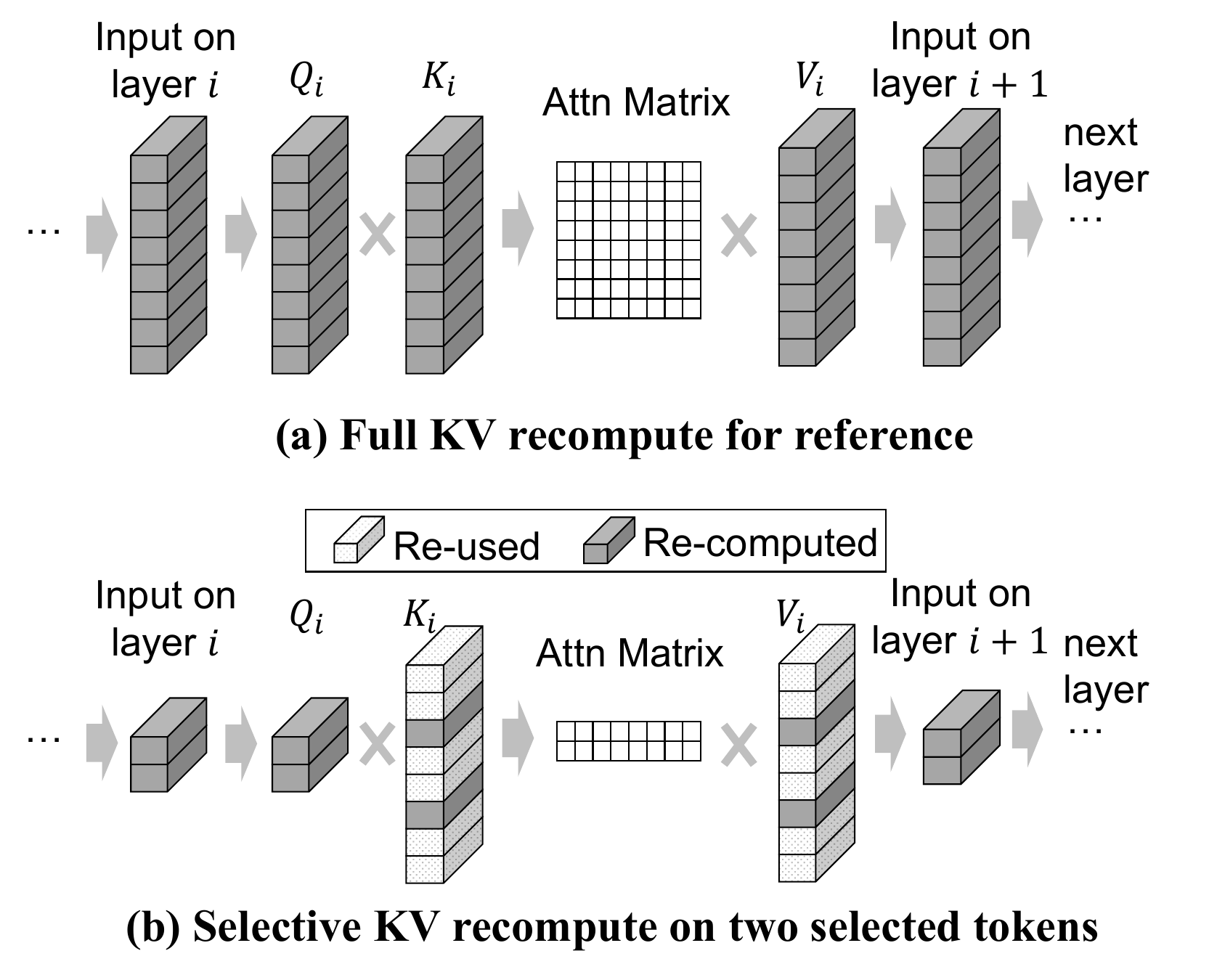}
    \tightcaption{Illustrated contrast between (a) full KV recompute and (b) selective KV recompute on one layer.}
    \label{fig:selective_recompute}
\end{figure}

For now, let us assume we have already selected a subset of tokens to recompute on each layer (we will explain how to select them in \S\ref{subsec:token-selection}). 
Here, we describe how \name recomputes the KV of these selected tokens on each layer. 

\mypara{Workflow} 
The default implementation of prefill (depicted in Figure~\ref{fig:selective_recompute}(a)) does not ``skip'' tokens while only computes the KV of a subset of tokens. 
Instead, \name runs the following steps (depicted in Figure~\ref{fig:selective_recompute}(b)):
\begin{packeditemize}
\item It first applies a mask on the input of each layer $\LayerIndex$ to reduce it to a subset of selected tokens.
\item It then transforms the reduced input into the $Q_{\LayerIndex}$, $K_{\LayerIndex}$ and $V_{\LayerIndex}$ vectors will also be restricted to the selected tokens. 
\item It then expands the $K_{\LayerIndex}$ vector and $V_{\LayerIndex}$ vector by reusing the KV cache entries associated with the un-selected tokens on layer $\LayerIndex$, so that the attention matrix includes attention between selected tokens and all other tokens.
\item Finally, it runs the same attention module to produce the input of the next layer. 
\end{packeditemize}
These changes make little assumption on the exact transformer process and can be integrated with many popular transformers (more details in \S\ref{sec:impl}). It is important to notice that the compute overhead is proportional to the number of selected tokens. 
This is because it only runs computation associated with the selected tokens. 
If we recompute $r\%$ of tokens per layer, the total compute overhead will be $r\%$ of full prefill. 

\subsection{Selecting which tokens to recompute}
\label{subsec:token-selection}

Next, we explain how to choose the tokens whose KV should be recomputed on each layer in order to reduce the attention deviation on each layer, which results from the KV deviation.
Our intuition, therefore, is to prioritize recomputing the KV of tokens who have high KV deviations.
Of course, this intuitive scheme is {\em not} feasible as it needs to know full-prefilled KV cache, and we will make it practical shortly.

\begin{figure}[t]
\centering
	\includegraphics[width=.74\columnwidth]{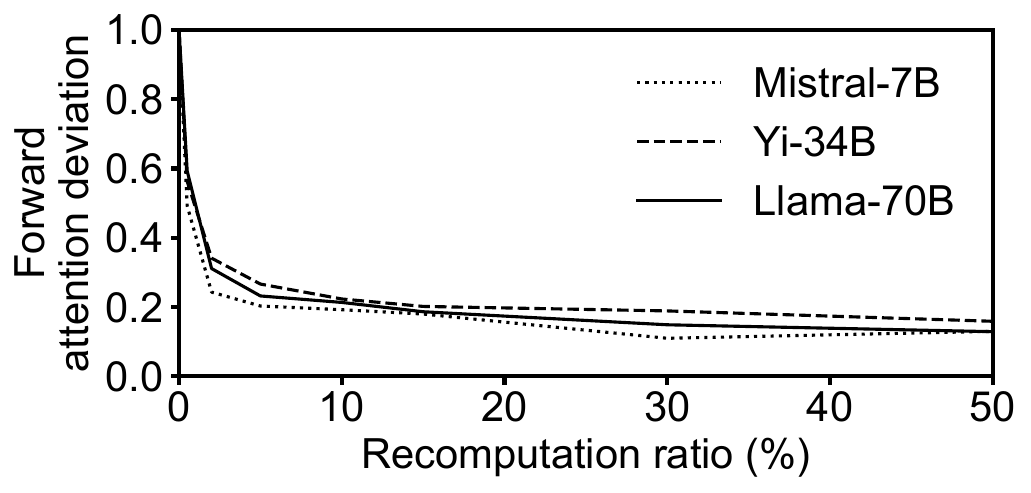}
    \tightcaption{Attention deviation reduces as we recompute the KV of more tokens on each layer. Importantly, the biggest drop in attention deviation results from recomputing the KV of the tokens with the highest KV deviation (\ie~\HCA tokens).}
    \label{fig:ca_reduction}
\end{figure}


To show the effectiveness of selecting tokens with high KV deviations, Figure~\ref{fig:ca_reduction} uses three models on the dataset of Musique (please see \S\ref{subsec:eval:setup} for details).
It shows the change of average attention deviation across all layers $\LayerIndex$, 
$\CAD(\Attention_{\LayerIndex}, \Attention_{\LayerIndex}^{\Full})$, after we use the aforementioned scheme (\S\ref{subsec:partial-recompute}) to select and recompute the $r\%$ of tokens $\TokenIndex$ who have the highest KV deviation 
$\KVD(\KV_{\LayerIndex}, \KV^{\Full}_{\LayerIndex})[\TokenIndex]$.
As the recompute ratio ($r$) increases, we can see that the attention deviation gradually reduces, and the biggest drops happen when the top few tokens with the highest KV deviations are recomputed. 
Empirically, it suggests the following insight.

\begin{insight}
On layer $\LayerIndex$, recomputing the KV of token $\TokenIndex$ who has a higher KV deviation (\ie $\KVD(\KV_{\LayerIndex}, \KV^{\Full}_{\LayerIndex})[\TokenIndex]$) reduces the attention deviation (\ie $\CAD(\Attention_{\LayerIndex}, \Attention_{\LayerIndex}^{\Full})$) by a greater amount. 
\end{insight}

Thus, if we recompute the KV of, say 10\%, of tokens on a layer $\LayerIndex$, we should choose the 10\% of tokens which have the highest KV deviations.\footnote{In the precomputed KV cache, the K vector of each chunk must be adjusted with the correct positional embedding. In SOTA positional embedding scheme (Rotary Positional Embedding or ROPE~\cite{su2024roformer}), this correction is done simply by multiplying the K vector by a rotation matrix of $\begin{pmatrix}
\cos m\theta & -\sin m\theta\\
\sin m\theta & \cos m\theta\\
\end{pmatrix}$. (The n-dimensional case in Appendix \ref{sec:appendix_pos}) This step has negligible overhead since the multiplication is performed only once.} 
We refer to these tokens as the {\bf High-KV-Deviation} (or {\bf \HCA}) tokens on layer $\LayerIndex$.

Now that we know we should recompute KV for the \HCA tokens, two natural questions arise.

\vspace{0.2cm}
\noindent {\bf Do we need to recompute KV for most tokens?}
In \S\ref{sec:eval}, we empirically show that choosing 10-20\% tokens as \HCA tokens and recomputing their KV suffices to greatly reduce the attention deviation and preserve generation quality.

\begin{figure}[t]
\centering
	\includegraphics[width=0.99\columnwidth]{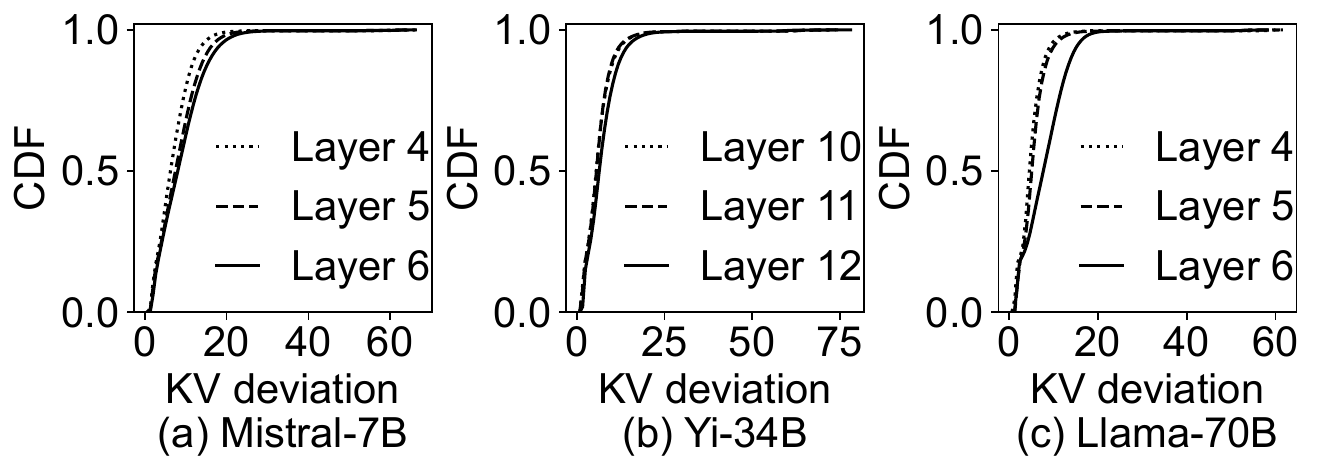}
    \tightcaption{Distribution of KV deviation of different tokens on one layer.}
    \label{fig:acacdf}
\end{figure}

This can be intuitively explained by {\em attention sparsity}, a well-studied property observed in many transformer models by prior research~\cite{liu2024scissorhands,h2o,choromanski2020rethinking,chen2021scatterbrain}. 
It says that in an attention matrix, high attention typically only occurs between a small number of tokens and their preceding tokens. 
To validate this observation, Figure~\ref{fig:acacdf} uses the same models and dataset as Figure~\ref{fig:ca_reduction}. 
It shows the distribution of KV deviation on one layer.
We can see that a small fraction, about 10-15\%, of tokens have much higher KV deviation than others, which corroborates the sparsity of cross-attention.

If a token has very low attention with other chunks' tokens (\ie low cross-attention with other chunks), the KV deviation between $\Attention^{\Pre}$ and $\Attention^{\Full}$ will be low and thus do not need to be recomputed.
Only when a token has a high attention with other chunks (high KV deviation compared with ground truth), should its KV be recomputed.

%

\vspace{0.2cm}
\noindent {\bf How to identify the \HCA tokens without knowing the true KV values or attention matrix?}
Naively, to identify the \HCA tokens, one must know the fully recomputed $\KV_{\LayerIndex}^{\Full}$ of each layer $\LayerIndex$ in the first place, but doing so is too expensive and defeats the purpose of selective KV recompute.
Instead, we observe that the \HCA tokens on different layers are not independent:

\begin{insight}
Tokens with the highest KV deviations on one layer are likely to have the highest KV deviations on the next layer. 
\end{insight}


\noindent For instance, if the \HCA tokens on the first layer are tokens 2, 3, and 5, these three tokens will likely also have higher KV deviations than most other tokens on the second layer. 

\begin{figure}[t]
\centering
	\includegraphics[width=.99\columnwidth]{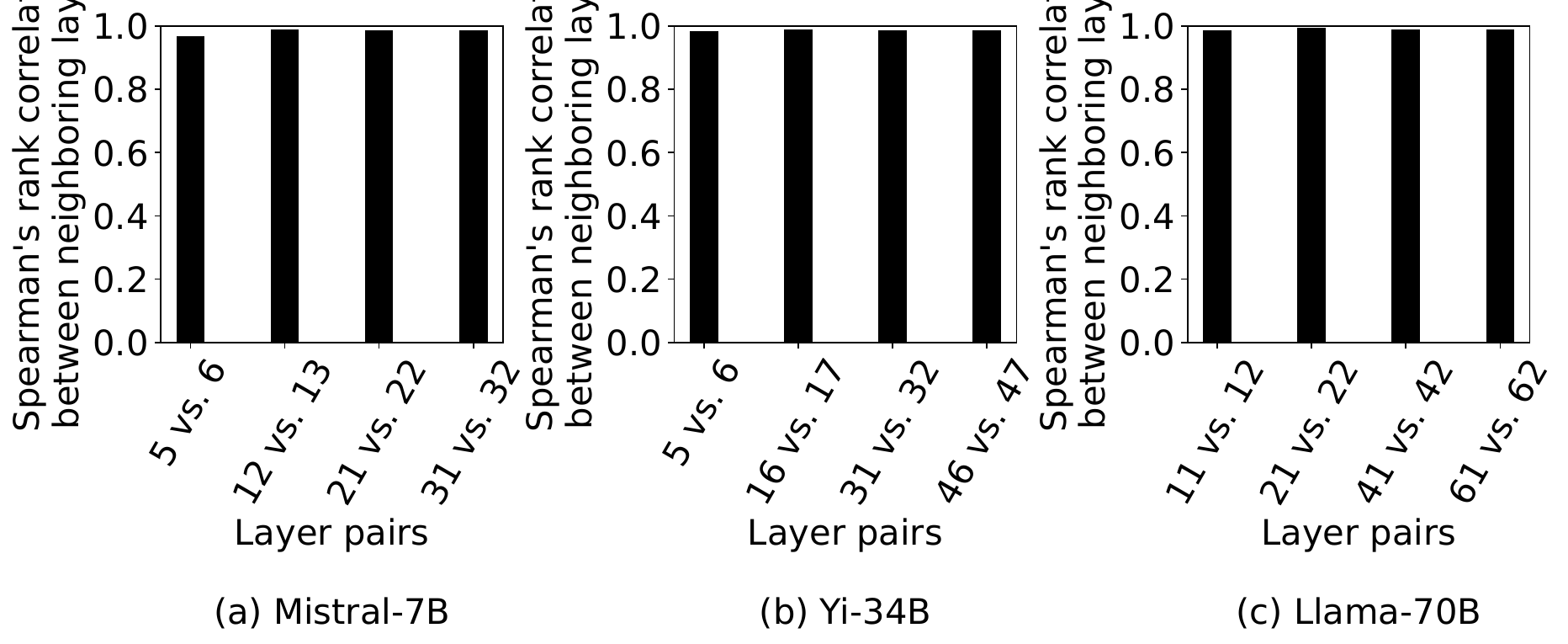}
    \tightcaption{Rank correlation of the KV deviation per token between two consecutive layers.}
    \label{fig:rankcorr}
\end{figure}

Figure~\ref{fig:rankcorr} uses the same setting as Figure~\ref{fig:acacdf} and shows Spearman's rank correlation score between the KV deviation of tokens between two neighboring layers. 
The figure shows a consistently high similarity of \HCA tokens between different layers.\footnote{We should clarify that although the \HCA tokens are similar across layers, the attention matrices between layers can still be quite different.} 

The intuition behind this correlation lies in the previous observation that the input embedding of each token changes slowly between layers in transformer models~\cite{liu2023deja,phang2021finesim}. Therefore, KV cache between layers should also bear similarity as KV cache is generated from the input embedding with a linear transformation.

Given the substantial correlation between the \HCA tokens, a straightforward solution is that we can perform prefill on the first layer first, pick the \HCA tokens of the first layer, and only update their KV on all other layers.
Since an LLM usually has over 30 layers, this process can save most of the compute compared to full KV recompute. 
That said, using {\em only} the attention deviation of different tokens on the first layer may not be statistically reliable to pick \HCA tokens of all layers, especially deeper layers. 

\begin{figure}[t]
\centering
	\includegraphics[width=.99\columnwidth]{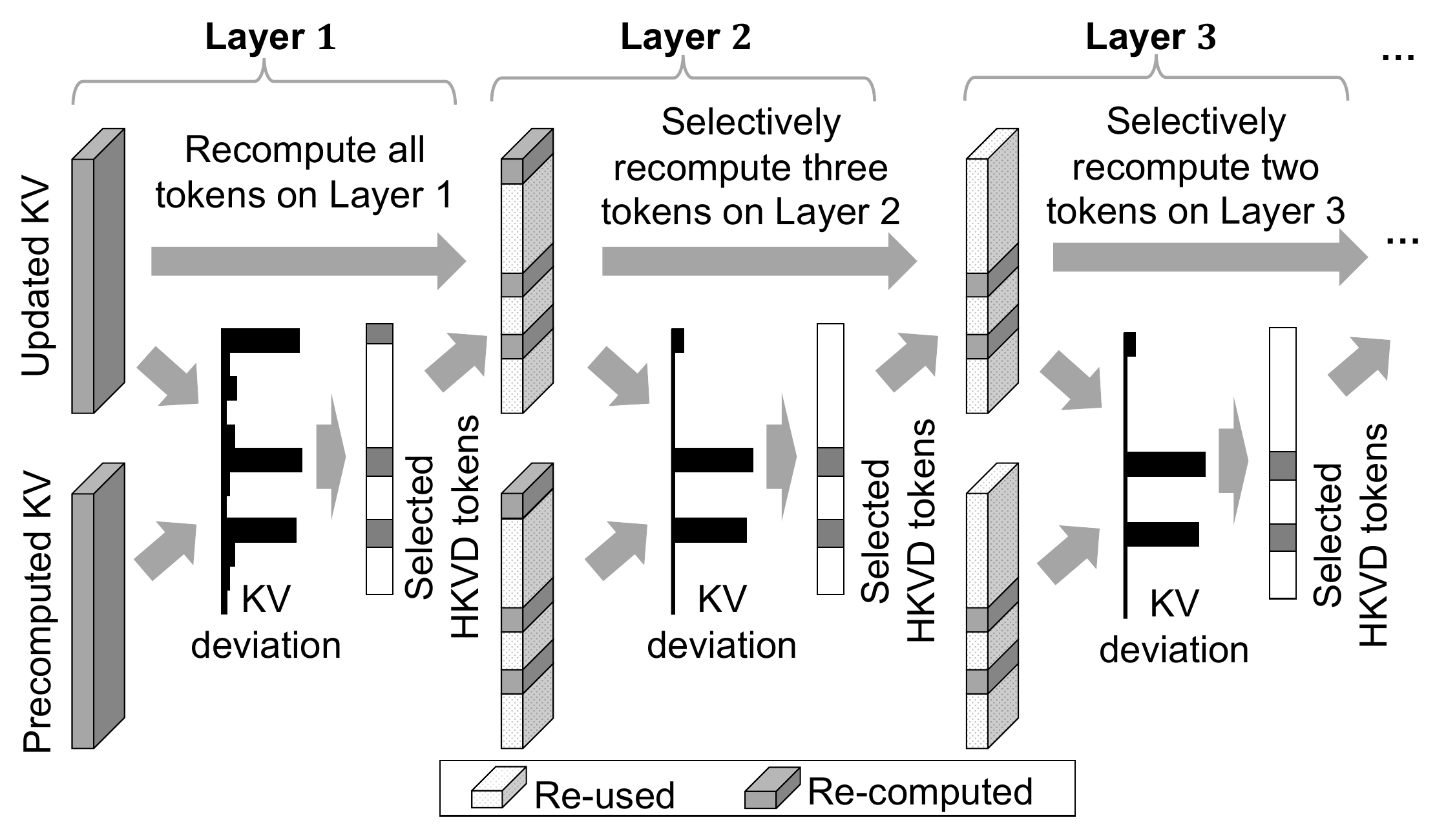}
    \tightcaption{\name selects the \HCA (high KV deviation) tokens of one layer by computing KV deviation of only the \HCA tokens selected from the previous layer and selecting the tokens among them with high KV deviation.}
    \label{fig:selection_illustrated}
\end{figure}

Thus, we opt for a {\em gradual filtering} scheme (depicted in Figure~\ref{fig:selection_illustrated}).
If on average we want to pick $r\%$ \HCA tokens per layer, we will pick $r_1\%$ tokens based on the token-wise attention deviation on the first layer, with $r_1$ being slightly higher than $r$, and use them as the \HCA tokens on the second layer.
Then we recompute the KV of these $r_1\%$ \HCA tokens on the second layer and pick $r_2\%$ tokens that have the highest token-wise attention deviation, with $r_2$ slightly less than $r_1$, as the \HCA tokens on the next layer, and so forth. 
Intuitively, this gradual-filtering scheme eventually picks the \HCA tokens who have high attention deviation, not only on the first layer but also on multiple layers, which empirically is statistically more reliable to identify the \HCA tokens on each layer.  

\hcedit{Although the KV-cache space of the layer i performing the HKVD calculation holds both Updated-KV and Precomputed-KV, layer-i’s extra Precomputed-KV is immediately discarded once the inference proceeds to layer i+1. This makes the memory overhead in HKVD negligible. }

\section{\name System Design}\label{sec:design}

We present a concrete system design for \name, which reduces the impact of the selective KV recompute using the following basic insight.

\vspace{0.2cm}
\mypara{Basic insight} \label{subsubsec:preemble}
\emph{If the delay for selective KV recompute (\S\ref{subsec:token-selection}) is faster than the loading of KV into GPU memory, then properly pipelining the selective KV recompute and KV loading makes the extra delay of KV recompute negligible.}
\vspace{0.2cm}


\mypara{Pipelining KV loading and recompute} 
In \name, the selective recompute of one layer can start immediately after pre-computed the KV cache of the previous layer is loaded into the GPU. 
This is because which tokens' KV to recompute on one layer only depends on the KV deviation of the previous layer's tokens. 
As a result, if loading the pre-computed KV for one layer is faster or equal to selective KV recompute of one layer, the KV-loading delay should be able to hide the selective recompute delay, \ie without incurring any extra delay on time-to-first-token (TTFT). 

Take the Llama-7B model and a 4K-long context, recomputing 15\% of the tokens (the default recompute ratio) only takes 3~ms per layer, while loading one layer's KV cache takes 16 ms from an NVME SSD (\S\ref{sec:eval}).
In this case, KV loading can hide the delay for KV recompute on 15\% of the tokens, \ie KV recompute incurs no extra delay.
Recomputing more tokens, which can slightly improve generation quality, may not incur extra delay either, as long as the delay is below 16~ms.
On the contrary, with another model, Llama-70B, recomputing 15\% of tokens takes 7~ms, but it only takes 4~ms to load one layer's KV from an NVME SSD.
Here KV loading does not completely hide the recompute delay. 
In short, a controller is needed to intelligently pick the recompute ratio as well as where to store the KV cache (if applicable). 

\subsection{Key Components}

To realize the benefit of pipelining KV loading and recompute, our system has three major components.

\begin{figure}
    \centering
    \hspace*{-0.35cm}\includegraphics[width=.92\columnwidth]{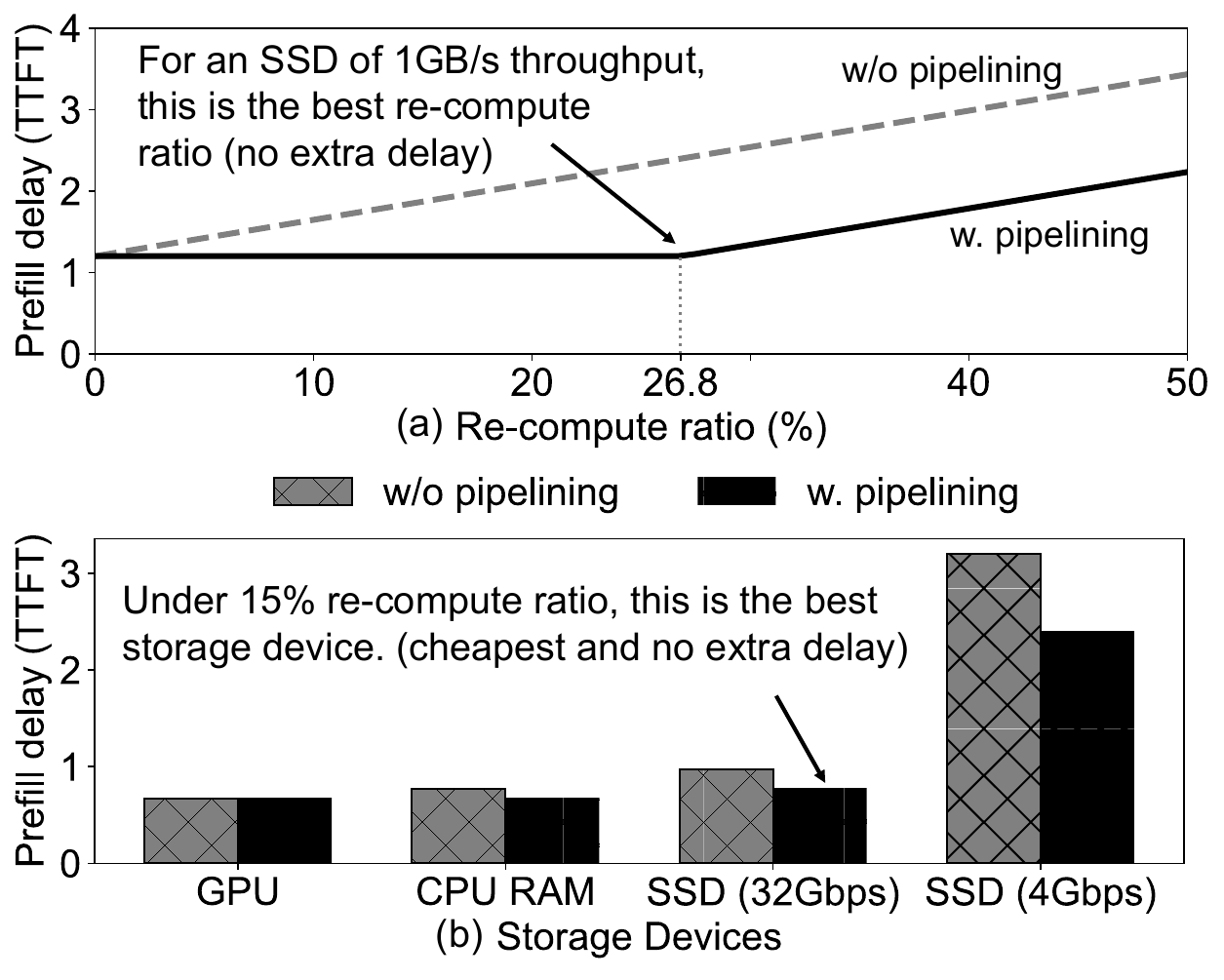}
    \vspace{-8pt}
    \tightcaption{(a) Smartly picking the recompute ratio will not incur an extra delay. (b) Smartly picking storage device(s) to store KVs saves cost while not increasing delay. }
    \label{fig:loading_controller}
\end{figure}

\begin{figure*}[h]
\centering
    \includegraphics[width=1.9\columnwidth]{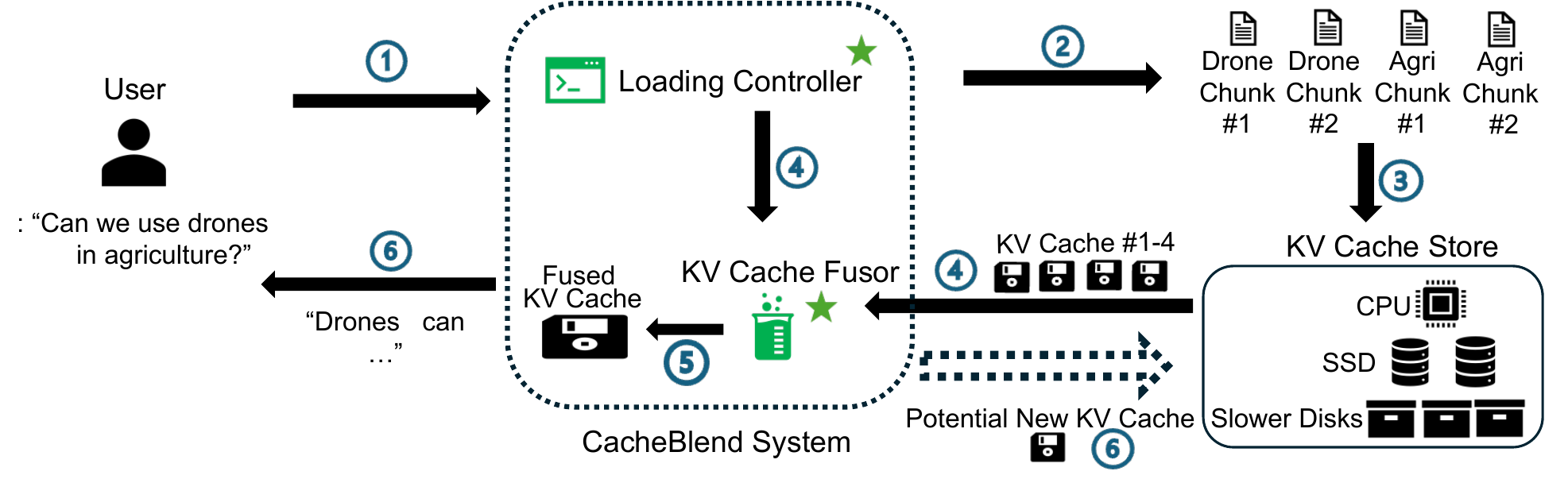}
    \tightcaption{\name system (green stared) in light of LLM context augmented generation for a single request. \name uses text provided by the retriever, interacts with  the storage device(s), and provides KV cache on top of LLM inference engines.}
    \label{fig:overview}
\end{figure*}

\mypara{Loading Controller} 
We face two design questions in practice: First, \emph{given a fixed storage device to use, how to choose a recompute ratio (what fraction of tokens to recompute KV per layer) without incurring extra delay to time-to-first-token (TTFT)? } 
Figure~\ref{fig:loading_controller}(a) illustrates an example that, if we select a recompute ratio wisely, the recompute should not cause \emph{any} extra delay to loading if loading is slow.

For this, the controller uses two delay estimators to find an idealized recompute ratio, such that the recompute delay is close to the loading delay. 
Given the recompute ratio $r$, length of context to be loaded $L$, and LLM, the {\em recompute delay estimator} calculates the expected delay $T_{recompute}(r\%, LLM, L)$\footnote{$T_{recompute}(r\%, LLM, L) = r\% \times Prefill(LLM, L)$. $Prefill(LLM, L)$ is offline profiled.}. 
The {\em loading delay estimator} estimates the loading delay of the KV cache of one layer, $T_{load}(LLM, L, storage\_device)$\footnote{$T_{load}(LLM, L, storage\_device) = \frac{PerTokenKVSize(LLM) \times L}{Throughput(storage\_device)}$.}, based on the LLM, the storage device's speed (which is measured offline), and the length of context $L$.

The controller calculates an idealized recomputation ratio such that the loading delay can hide the recompute delay, without degrading the inference quality. 
It first picks the recompute ratio $r\%$ such that $T_{recompute}(r\%, LLM, L)$ is equal to $T_{load}(LLM, L, storage\_device)$, and then takes the max of $r\%$ and $r^*\%$, where $r^*\%$ is the minimal recompute ratio that empirically has low negligible quality drop from full KV recompute. 
In practice, we found $r^*\%$ to be 15\% from Figure~\ref{fig:overall_recomp}. \hcedit{This means that even if the storage device is a fast device (ex. CPU RAM), the delay will be lower-bounded by the minimal recomputation to guarantee quality.}

\hcedit{In practice, \name faces another challenge: which {\em storage devices} should the developer use? To solve this challenge, we present a more formulated question to the loading controller:} \emph{If we only do KV recompute of a fixed selective recompute ratio (ex. $15\%$), how can we choose the right storage device to store KVs such that no extra delay is caused?} As shown in Figure~\ref{fig:loading_controller}(b), under a fixed recompute ratio, the controller should pick the cheapest storage device among all devices that do not increase the delay.

In \name, the system developers can provide a list of potential storage device(s), and the controller
uses a {\em storage cost estimator} which estimates the cost of storing KVs for each device, namely $C_{store}(LLM, L, T, storage\_device)$, based on the LLM, length of context $L$ and time duration $T$ needed to store it (if it is cloud storage).
Then it uses $T_{recompute}(15\%, LLM, L)$ and $T_{load}(LLM, L, storage\_device)$ to estimate the recompute and loading delays for all devices. 
Lastly, it finds out which storage device is the cheapest where $T_{recompute}\geq T_{load}$.
\hcedit{In this way, if the developer can navigate through the different storage devices for the KV caches given a fixed recomputation target that satisfies the generation quality requirement.}


\mypara{KV cache store {\em (mapping LLM input to KV caches)}} 
The KV cache store splits an LLM input into multiple text chunks, each of which can be reused or new. 
For instance, a RAG input typically consists of multiple retrieved context chunks (likely of a fixed length) and the user input. 
The splitting of LLM inputs is specific to the application, and we implement the same strategy as described in recent work~\cite{gim2023prompt,parrot}. 
Once the input is split into text chunks, each chunk is hashed to find their corresponding KV cache, in the same way as the block hashing is  implemented in vLLM~\cite{kwon2023efficient}. 
The KV caches of new chunks generated by the fusor (explained soon) are added to the devices. When the storage devices are full, we evict the least recently used KV cache. \hcedit{In this paper, we only focus on storing KV cache in one single level of storage device such as CPU RAM or SSD.}

\mypara{Fusor} The cache fusor (\S\ref{sec:merge}) merges pre-computed KV caches via selective recompute. 
Recall from \S\ref{subsec:token-selection}, the decision of which tokens need to be recomputed for one layer depends on the recompute of the previous layer.
Thus, the fusor waits until the recompute for the previous layer is done, and the KV caches for layer $L$ are loaded into the queue on GPU memory and then perform selective recompute using the recompute ratio $r\%$ calculated by the loading controller. 
The fusor repeats this process until all the layers are recomputed. 


\subsection{Putting them together}
We  put the key components together in an LLM inference workflow in Figure~\ref{fig:overview}. 
When a user of an LLM application submits a question, a list of relevant text chunks will be queried. 
The loading controller then queries the KV cache manager on whether the KV caches for those text chunks exist, and where they are stored. 
Next, the KV cache manager returns this information back to the loading controller and the controller computes the idealized selective recomputation ratio, sends it to the fusor, and loads the KV caches into a queue in GPU memory. 
The KV cache fusor continuously recomputes the KV caches in the queue, until all layers are recomputed. 
Lastly, the fused KV cache is input into the LLM inference engine, which generates the answer to the user question based on the KV cache.

\section{Implementation}\label{sec:impl}
We implement \name on top of vLLM with about 3K lines of code in Python based on PyTorch v2.0.

\mypara{Integrating Fusor into LLM serving engine}
\name performs the partial prefill process in a layer-wise manner through three interfaces:

\begin{packeditemize}
\item \texttt{fetch\_kv(text, layer\_id) -> KVCache}: given a piece of text and a layer id, \name fetches the corresponding KV cache from KV store into the GPU. Returns -1 if the KV cache is not in the system.
\item \texttt{prefill\_layer(input\_dict, KVCache) -> output\_dict}: \name takes in the input and KV cache of this layer and performs the partial prefill process for this particular layer. The output is used as the input for the next layer.
\item \texttt{synchronize()}: \name requires synchronization before prefilling every layer to make sure the KV cache of this layer has already been loaded into the GPU.
\end{packeditemize}

We implement these three interfaces inside vLLMs. For \texttt{fetch\_kv}, we first calculate the hash of the text and search if it is inside the KV store system. If it is present, we call \texttt{torch.load()} to load it into GPU memory if KV cache is on disk or use \texttt{torch.cuda()} if the KV cache is inside CPU memory. For \texttt{prefill\_layer}, we implement this interface on top of the original \texttt{layer} function in vLLM that performs one layer of prefill. Three key-value pairs are recorded in the \texttt{input\_dict}: (1) the original input data \texttt{input\_org} required for prefilling an LLM layer (e.g., input\_tensor, input\_metadata), (2) a \texttt{check\_flag} indicating whether \HCA tokens will be selected in this layer, and (3) \texttt{HKVD\_indices} that track the indices of \HCA tokens. If \texttt{check\_flag} is \texttt{True}, the input tokens with the largest deviation between the newly computed KV cache and the loaded KV cache will be selected as the \HCA tokens. If \texttt{check\_flag} is \texttt{False}, partial prefill will only be performed on the current \HCA tokens indicated by \texttt{HKVD\_indices}. Only the KV cache of the \HCA tokens will be computed and updated at each layer.
In the partial prefill for layer \(i\), two threads are used to pipeline the computation (\texttt{prefill\_layer}) of layer \(i\) and the KV cache loading (\texttt{fetch\_kv}) of the next layer \(i+1\). \texttt{synchronize} is called before \texttt{prefill\_layer} to assure the KV cache needed for prefill has been loaded into GPU.



\mypara{Managing KV cache}
\name manages the KV caches such that:
If KV cache is not inside the system and is recomputed by the LLM engine in the runtime, we will move the KV cache into CPU by \texttt{torch.cpu()} and open a thread to write it back to disk in the background with \texttt{torch.save()}. During \texttt{fetch\_kv}, we go through through the hash tables to fetch KV cache for the fusor. The hash tables are kept in CPU for their relatively small size (16MB for one million chunks).



\vspace{-5pt}

\begin{figure*}[ht]
\centering
    \hspace*{-0.6cm}\includegraphics[width=1.8\columnwidth]{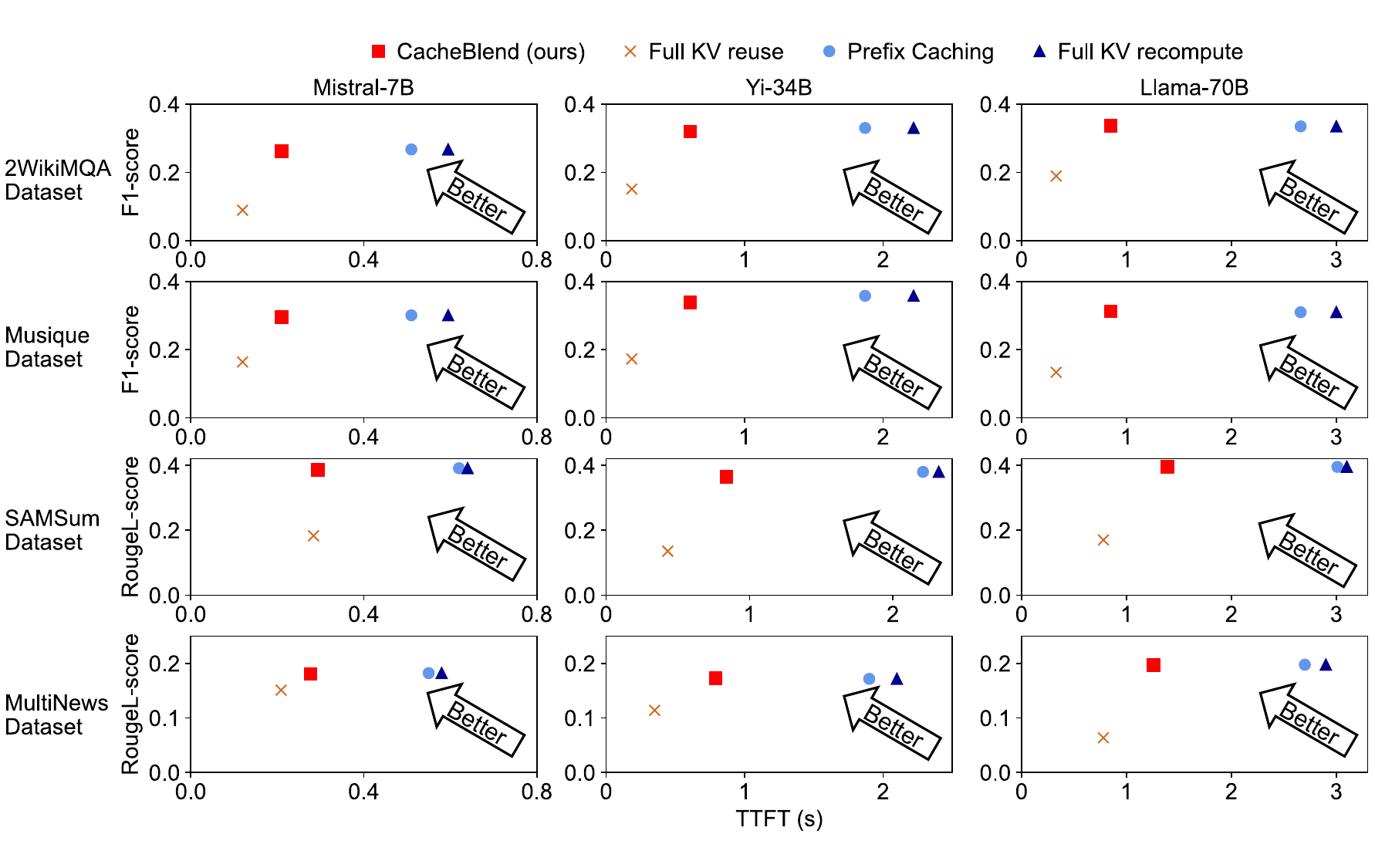}

    \vspace{-5pt}
    \tightcaption{\name reduces TTFT by 2.2-3.3$\times$ compared to full KV recompute with negligible quality drop across four datasets and three models. }
    \label{fig:e2e_delay_vs_tp}
\end{figure*}

\begin{figure*}[ht]
 \centering
     \hspace*{-0.4cm}\includegraphics[width=2\columnwidth]{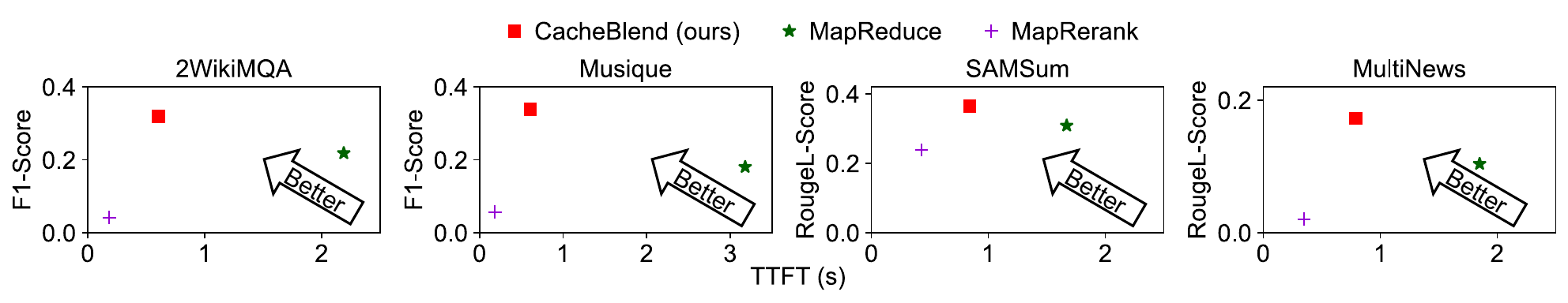}
    \tightcaption{Generation quality of \name with Yi-34B vs MapReduce and MapRerank. }
    \label{fig:map_reduce_rerank}
 \end{figure*}

\section{Evaluation}\label{sec:eval}
\noindent
Our key takeaways from the evaluation are: 
\begin{itemize}
    
    \item \textbf{TTFT reduction:} Compared to full KV recompute, \newline \name reduces TTFT by 2.2-3.3$\times$ over several models and tasks.
    \item \textbf{High quality:} Compared with full KV reuse, \name improves quality from 0.15 to 0.35 in F1-score and Rouge-L score, while having no more than 0.01-0.03 quality drop compared to full KV recompute and prefix caching. 
    \item \textbf{Higher throughput:} At the same TTFT, \name can increase throughput by up to 5$\times$ compared with full KV recompute and 3.3$\times$ compared with prefix caching. 


\end{itemize}
\begin{figure*}[ht]
\centering
    \hspace{-0.3cm}\includegraphics[width=1.8\columnwidth]{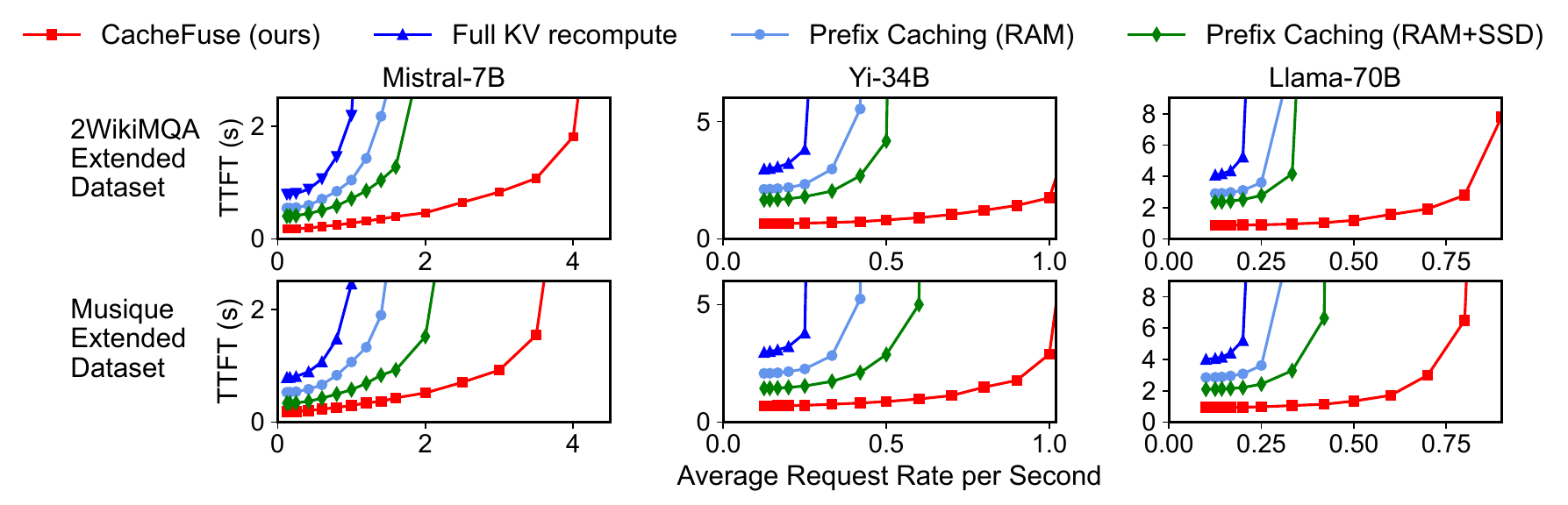}
    \tightcaption{\name achieves lower TTFT with higher throughput in RAG scenarios compared with baselines of similar quality.}
    \label{fig:e2e_delay_vs_request_rate}
\end{figure*}

\begin{figure}
\centering

    \includegraphics[width=1.0\columnwidth]{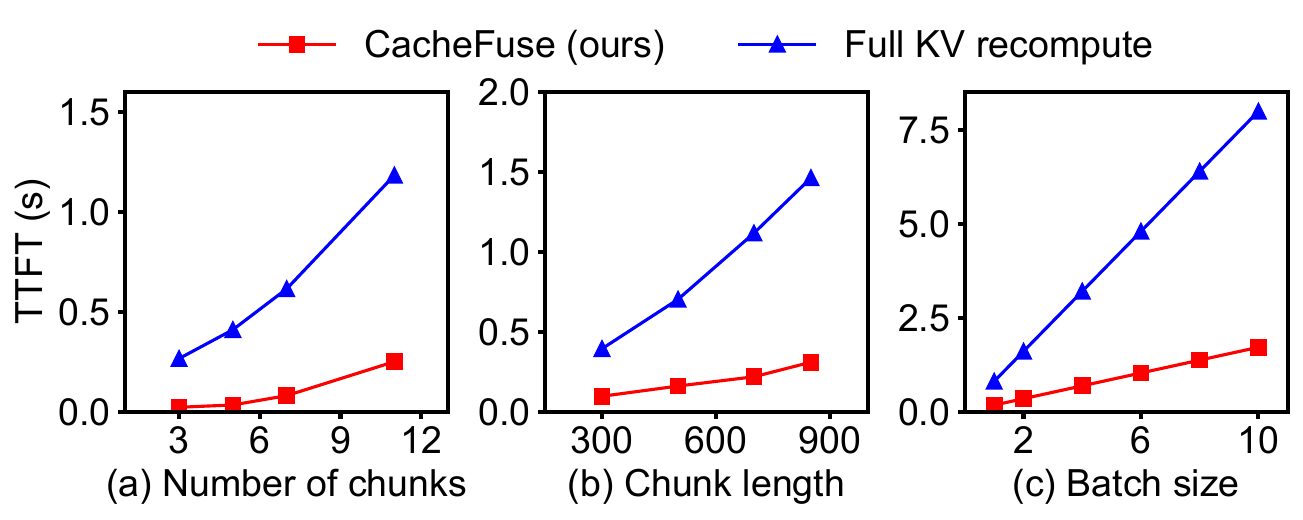}
    \vspace{-10pt}
    \tightcaption{\name outperforms baseline with varying chunk numbers, chunk lengths, and batch sizes. }
    \label{fig:chunk_batch}
\end{figure}
\subsection{Setup}\label{subsec:eval:setup}

\mypara{Models and hardware settings} We evaluate \name on Mistral-7B\cite{jiang2023mistral}, Yi-34B\cite{young2024yi} and Llama-70B\cite{llama} to represent a wide scale of open source models. Note that we apply 8-bit model quantization to Llama-70B and Yi-34B. 
We run our end-to-end experiments on Runpod GPUs~\cite{runpod2024} with 128~GB RAM, 2 Nvidia A40 GPUs, and 1TB NVME SSD whose measured throughput is 4.8~GB/s. 
We use 1 GPU to serve Mistral-7B and Yi-34B, and 2 GPUs to serve Llama-70B.

\mypara{Datasets} Our evaluation covers the following datasets.  
\label{subsec:dataset}
\begin{packeditemize}
    \item \textit{2WikiMQA}\footnote{
            Since the standard answers for \textit{2WikiMQA} and \textit{Musique} are always less than 5 words,
            we append ``Answer within 5 words.'' to their prompts to reduce the impact of answer length mismatch in F1 score calculation.
            \label{note_dataset}}~\cite{2wikimqa}: This dataset aims to test LLM's reasoning skills by requiring the model to read multiple paragraphs to answer a given question. We included 200 test cases, following the dataset size of previous work~\cite{longbench}. 
    \item \textit{Musique}\footnotemark[\getrefnumber{note_dataset}]~\cite{trivedi2022musique}: This is a multi-document question-answering dataset. It is designated to test LLM's multi-hop reasoning ability where one reasoning step critically relies on information from another and contains 150 test cases.
    \item \textit{SAMSum}~\cite{samsum}: This dataset comprises multiple pairs of dialogues and summaries, and requires the LLM to output a summary to a new dialogue. It is intended to test the few-shot learning ability of language models and contains 200 test cases.
    \item \textit{MultiNews}~\cite{multinews}: This dataset consists of news articles and human-written summaries of these articles from the site newser.com. Each summary is professionally written by editors and includes links to the original articles cited and contains 60 sampled cases.
    
    
\end{packeditemize}
\hcedit{We split contexts into 512-token chunks with Langchain and use the original 200-400 token chunks in SAMSum.}
We also create a synthetic dataset to simulate the chunk reuse in RAG scenarios. 
Specifically,
we randomly pick  1500 queries in the Musique and 2WikiMQA datasets each and build a context chunk database by splitting each query's context into 512-token chunks~\cite{jan_chunksize} with Langchain~\cite{langchain}. For each query, we use GPT4 API to generate 3 more similar queries. In the 6000 queries (1500 original + 4500 simulated), we retrieve the top-6 chunks\footnote{Max number of chunks that fit into input token limit for Llama-70B. } based on L2 distance, in a random order\cite{johnson2017billionscale}. 
We refer to these datasets as \emph{Musique extended} and \emph{2WikiMQA extended}. 
We only report for baselines with similar quality and skip the result for the first 1K queries as the initial storage is completely empty.

\vspace{\vspacesize}
\noindent

\mypara{Quality metrics} We adopt the following standard metrics to measure the generation quality.
\begin{packeditemize}
    \item \textit{F1-score~\cite{f1}} is used to evaluate \textit{2WikiMQA} and \textit{Musique} datasets~\cite{longbench}. It measures the similarity between the model's output and the ground-truth answer of the question based on the number of overlapping words.
    \item \textit{Rouge-L score~\cite{lin-2004-rouge}} is used to evaluate \textit{MultiNews} and \textit{SAMSum} datasets~\cite{longbench}. It measures the similarity between the model's output and the ground-truth summaries based on the longest common sequence.
\end{packeditemize}



\mypara{Baselines} We compare \name with the following baselines:
\begin{itemize}
    \item \textit{Full KV recompute}: The raw texts are fed into LLM as input. The LLM calculates KV cache of all tokens during prefill. 
    \item \textit{Prefix caching}~\cite{jin2024ragcache, sglang, kwon2023efficient}: 
    We adopt the techniques from SGLang~\cite{sglang} to identify the frequently used prefix chunks and store their KV caches in both RAM and SSD.
    The KV cache of non-prefix tokens needs to be computed during prefill. 
    We also make an idealized assumption in favor of prefix caching that there is no loading delay from RAM or SSD to GPU. 
    This assumption makes it perform better than it would under real-world conditions.
    \item \textit{Full KV reuse}~\cite{gim2023prompt}: We implement full KV reuse by using the approach proposed in \promptcache~\cite{gim2023prompt}. \hcedit{We append a buffer before the text to prepare its KV cache to be used in different positions with correct positional encoding. We did not compare with the scaffolding scheme since its application to RAG scenarios requires human users to manually select important chunks at runtime.}
    \item \textit{MapReduce}~\cite{mapreduce-langchain}: Different from traditional MapReduce~\cite{mapreduce}, this is an alternative RAG method in LangChain. The LLM first summarises all chunks in parallel and concatenates them together.
    The concatenated summaries are then fed to the LLM again to generate the final answer. 
    \item \textit{MapRerank}~\cite{maprerank-langchain}: This is another RAG method in LangChain. In MapRerank, the LLM independently generates an answer from each chunk along with a score based on its confidence that the answer is correct. The answer with the highest score is picked as the final output.

\end{itemize}

 \begin{figure*}[ht]
 \centering
     \hspace*{-0.4cm}\includegraphics[width=2\columnwidth]{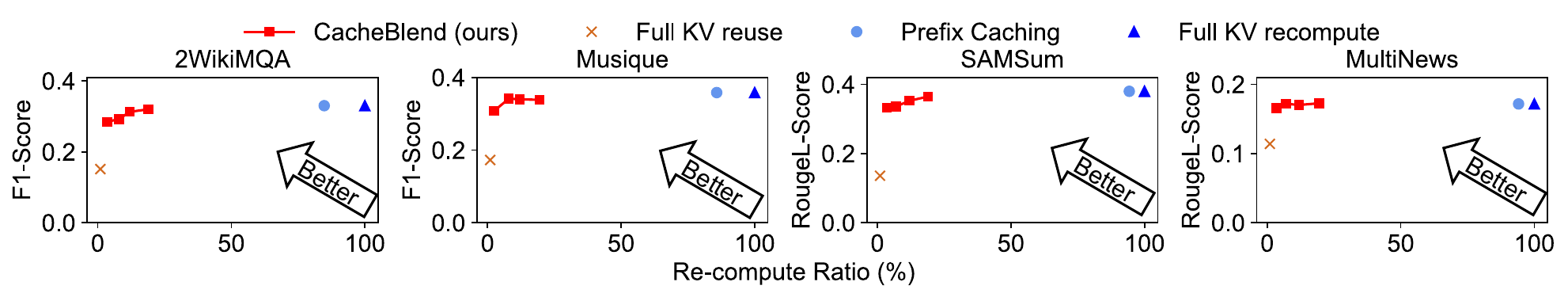}
    \tightcaption{\name has minimal loss in quality compared with full KV recompute, with 5\%--18\% selective recompute ratio, with Yi-34B.
    }
    \label{fig:overall_recomp}
 \end{figure*}
 
\begin{figure}[ht]
\centering
\hspace*{-0.5cm} \includegraphics[width=1.1\columnwidth]{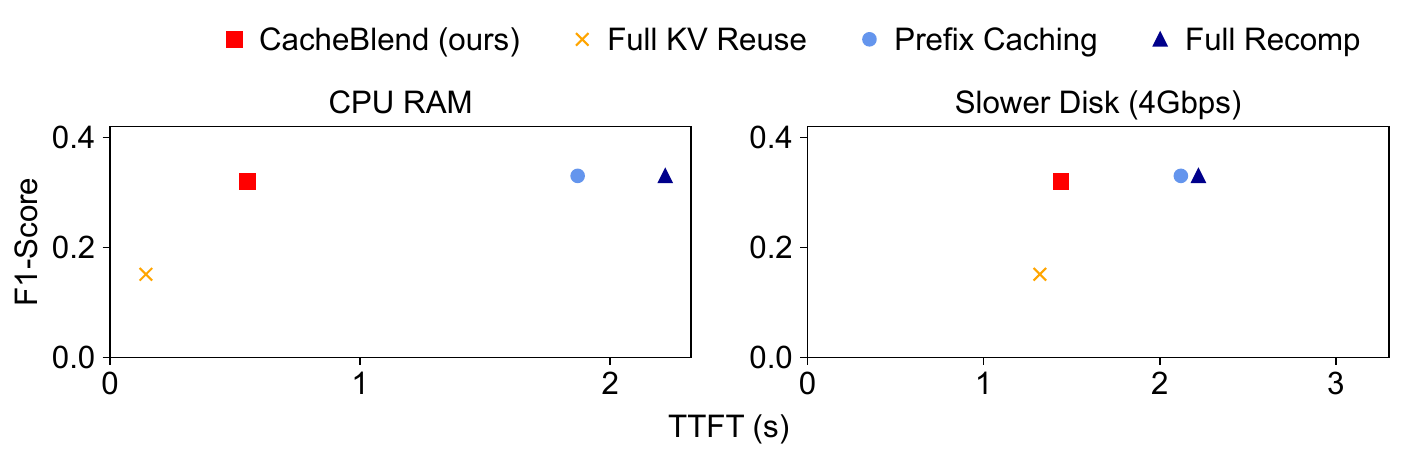}
    \vspace{-10pt}
    \tightcaption{\name's outperforms baselines when using RAM and slower disks}
    \label{fig:storage_device}
\end{figure}

\subsection{Overall Improvement}\label{overall improve}

\mypara{Reduced TTFT with minimal quality drop}
Figure~\ref{fig:e2e_delay_vs_tp} compares the average quality and TTFT across the requests, where each request has a context of 6 top chunks picked by lowest L2-distance between the respective embeddings generated by SentenceTransformers~\cite{reimers-2019-sentence-bert} (512 tokens per chunk).
As shown in the graph, compared to the full KV recompute and prefix caching, \name's reduction in F1 and Rouge-L score is within 0.02, while it significantly reduces the TTFT by 2.2-3.3$\times$ across all models and datasets. While \name is slower than full KV reuse due to its selective recomputation, its quality stably outperforms full KV reuse by a large margin (in many cases more than 2$\times$). 

Figure~\ref{fig:map_reduce_rerank} compares \name with RAG methods including MapReduce and MapRerank.
Compared to MapReduce, \name has a 2-5$\times$ lower TTFT and higher F1 score. 




\mypara{Higher throughput with lower delay}
\label{subsec:overall_improvement}
In Figure \ref{fig:e2e_delay_vs_request_rate}, we compare \name with full KV recompute and prefix caching on Musique extended and 2WikiMQA datasets under different request rates. \name achieves lower delay with higher throughput by 2.8-5$\times$ than all the baselines across different models and datasets.

\mypara{Understanding \name's improvement} \name is better than all baselines for different reasons. 
Compared to the full KV recompute, \name has a much lower delay and higher throughput due to only a small amount of tokens are recomputed. 
Compared to full KV reuse, although its delay is lower than \name, the quality drops a lot as full KV reuse did not perform any of the recompute, thus missing the cross-attention between different chunks. 
Compared to prefix caching, \name is also better in terms of higher throughput and lower delay as prefix caching needs to store \emph{multiple versions} of KV caches for the same chunk if they have different prefixes. 
Thus,
given the total storage space is fixed, prefix caching will incur a higher miss rate. 

Finally, compared to other RAG methods, like MapReduce and MapRerank, \name is also better in terms of quality or delay. For MapReduce, it has a higher delay than \name due to additional LLM inference. 
Although MapRerank has slightly lower TTFT than \name, its quality is much worse, since processing the input chunks separately ignores the dependencies between chunks.

\subsection{Sensitivity Analysis}
\label{subsec:sensitivity}

For a better understanding of \name, we further analyze how varying the configurations impacts overall performance.

\mypara{Varying chunk numbers and lengths} 
Figure \ref{fig:chunk_batch}a and \ref{fig:chunk_batch}b show the minimum compute time needed by \name to maintain generation quality ($\leq$0.015 loss in F1-score) at different numbers of chunks and chunk lengths. 
The experiment is conducted on 2WikiMQA with Mistral-7B model. 
As shown in the figure, the compute time reduction ratio remains similar across different numbers of chunks and chunk length settings.

\mypara{Varying recompute ratios} 
Figure \ref{fig:overall_recomp} shows the impact of the recompute ratio on the quality-TTFT trade-off across all datasets on Yi-34B model.
Across all the datasets, \name's loss in generation quality is at most 0.002 in F1 score or Rouge-L score compared to full KV recompute,
with 5\%\textasciitilde18\% recomputation ratio. 
To put the number into context, the 5\%\textasciitilde18\% recomputation ratio can be translated to 4.1-6.6$\times$ TTFT reduction compared with full KV-recompute and 3.4-6.1$\times$ TTFT reduction compared with prefix caching.

\mypara{Varying batch size}
Figure~\ref{fig:chunk_batch}c shows the compute time of prefill phase of different batch sizes. 
It is worth noting that the time of the decoding phase increases slower than the prefill phase when the batch size becomes larger~\cite{zhong2024distserve,kwon2023efficient}, making prefill overhead dominant with increasing batch size.
Therefore, \name's improvement over the prefill phase becomes more prominent to the overall delay reduction with larger batch sizes.

\mypara{Varying storage device}
To study the effect of different storage types on \name, we modify the underlying storage devices for every method and conduct a similar experiment as Figure \ref{fig:e2e_delay_vs_tp} for the Yi-34B model and 2WikiMQA dataset. 
As shown in Figure~\ref{fig:storage_device}, \name consistently reduces TTFT with minimal quality degradation when the KV cache is stored in RAM or a slower SSD device. \hcedit{Notice that the delay gap between \name and Full KV reuse is smaller for slower storage since the delay of \name would be more dominated by the loading delay instead of its partial KV recomputation.}

\section{Related Work}

\vspace{\vspacesize}
\noindent\mypara{Retrieval augmented generation (RAG)}
RAG \cite{mao2020generation, ram-etal-2023-context, gao2023retrievalaugmented, li2022ragsurvey, gao2023ragsurvey} can enhance the accuracy and reliability of LLMs with text chunks fetched from external sources. However, processing these text chunks in the LLM can take a long time. \name reduces this overhead by storing and reusing the KV caches of these text chunks.

\vspace{\vspacesize}
\noindent\mypara{KV cache reuse across requests} 
Storing and reusing KV cache across different requests have been commonly studied in recent work \cite{sglang, cachegen, gim2023prompt, liu2024optimizingsql, jin2024ragcache}. Most of these works \cite{sglang, cachegen, liu2024optimizingsql, jin2024ragcache} focus on prefix-only caching. \promptcache \cite{gim2023prompt} allows KV cache to be reused at different positions but fails to maintain satisfying generation quality due to inaccurate postional encoding and ignorance of cross attention. \name adopts a novel partial recomputation framework to better retain positional accuracies and cross attention. Most of the existing work stores KV cache in volatile memory devices for guaranteed performance (e.g., GPU HBM, CPU DRAM). While there are emerging research trying to reuse high-speed NVME SSD for KV caches~\cite{gao2024attentionstore}, \name is unique in pipelining loading with partial recomputation and its extension to even slower object store.

\vspace{\vspacesize}
\noindent\mypara{General-purpose LLM serving systems}
Numerous general-purpose LLM serving systems have been developed \cite{orca, kwon2023efficient, agrawal2023sarathi, zhong2024distserve}. Orca \cite{orca} enables multiple requests to be processed in parallel with iteration-level scheduling. vLLM \cite{kwon2023efficient} further increases the parallelsim through more efficent GPU memory management.
\name is complementary to these general-purpose LLM serving systems, empowering them with context resuing capabilities. 

\vspace{\vspacesize}
\noindent\mypara{Context compression methods}
Context compression techniques \cite{h2o, jiang2023llmlingua,jiang2023longllmlingua, liu2024scissorhands, dong2024getmoreless,yin2024llm} can be complementary to \name. Some of these techniques \cite{jiang2023llmlingua,jiang2023longllmlingua} shorten the prompt length by prunining the unimportant tokens. \name is compatible with such methods in that it can take different chunk lengths as shown in \S\ref{subsec:sensitivity}. Another line of work \cite{h2o, liu2024scissorhands,dong2024getmoreless} focus on dropping the unimportant KV vectors based on the attention matrix, which essentially reduce the KV cache size. \name can benifit from such techniques by storing and loading less KV cache.


\section{Limitations and Future Work}

Our method in this paper (e.g., the insights in \S~\ref{subsec:token-selection}) currently only applies to language models with transformer structures. We leave investigation of architectures other than transformer such as Mamba~\cite{gu2023mamba} and Griffin~\cite{de2024griffin} for future work.
In our evaluation, \hcedit{we haven't tested \name's performance on more models and datasets with different quantization settings.} For better understanding and improvement of the method, we have open-sourced our work to facilitate more effort in this direction.

In this paper, we integrated \name in vLLM but  have not yet tested \name's performance on the latest serving engines like Distserve\cite{zhong2024distserve} or StableGen
\cite{agrawal2023sarathi}. \hcedit{Nor have we studied how to apply \name to workloads that share KV cache across different compute nodes.} Since \name is able to reduce the costly prefill phase, we believe combining \name with these new serving engines could potentially bring more savings. We leave the integration of \name into these novel inference frameworks for future work. 



\section{Conclusion}
We present \name, a system that combines multiple pre-computed KV caches, when their corresponding texts are concatenated in the LLM input. 
To preserve generation quality, \name recovers the cross-attention among these texts  by selectively recomputing the KV cache values of a small fraction of tokens.
Through experiments across four datasets and three models, \name reduces TTFT by 2.2-3.3$\times$ and increases throughput by 2.8-5$\times$, compared to full KV recompute, under negligible quality drop. The code is available at \href{https://github.com/LMCache/LMCache}{\textcolor{blue}{\textbf{https://github.com/LMCache/LMCache}}}.

\section{Acknowledgement}
We thank all the anonymous reviewers and our shepherd, Thaleia Dimitra Doudali, for their insightful feedback and suggestions. The project is
funded by NSF CNS-2146496, CNS-2131826, CNS-2313190, CNS-1901466, CNS-2313190, CCF-2119184, CNS-1956180, and research awards from Google, CERES Center, Conviva, and Chameleon Cloud.

\bibliographystyle{plain}
\bibliography{citations} 

\pagebreak
\appendix
\label{sec:appendix}

\section{N-dimensional positional recovery}\label{sec:appendix_pos}
Here we prove our positional recovery method can work in the N-dimensional scenario. We start with the definition of RoPE in N-dimensional space.

\begin{definition}[Rotary Positional Encoding, ROPE\cite{su2024roformer}]
Let vectors \( {q}, {k}\in\mathbb{R}^{d} \) denote the query vector and key vector need to be embedded at some position m as \({q}_{m}, {k}_{m}\in\mathbb{R}^{d}\). Rope encodes the positional information as the following:

\[ {{q}_{m}, {k}_{m}}=\mathbb{R}^{d}_{\Theta, m} \{q,k\}\]
where

\footnotesize{
\[ \mathbb{R}^{d}_{\Theta, m}= 
\begin{pmatrix}
\cos m\theta_{0} & -\sin m\theta_{0} & ... & 0 & 0\\
\sin m\theta_{0} & \cos m\theta_{0}  & ... & 0 & 0\\
\vdots & \vdots & \ddots & \vdots &\vdots\\
0 & 0 &\vdots & \cos m\theta_{\frac{d}{2}-1} & -\sin m\theta_{\frac{d}{2}-1}\\
0 & 0 &\vdots & \sin m\theta_{\frac{d}{2}-1} & \cos m\theta_{\frac{d}{2}-1} \\
\end{pmatrix}\]
}

is the rotary matrix with hyperparameter \(\Theta\in \{ \theta_{i} = 10000^{-2id}, i \in [0,1,...,\frac{d}{2}-1]\}\)

\end{definition}

The reason why our positional recovery method can work is because attention score between a pair of tokens is invaraint to their absolute positions. Below is the proof of this invariance.

\begin{proposition}[Rope only depends on relative position]
Let vector \( {k}\in\mathbb{R}^{d} \) denote a key vector and \( {k}_{m}\in\mathbb{R}^{d} \) denote the key vector embedded at the fixed position m. And let vector \({q}\in\mathbb{R}^{d} \) denote a query vector and \( {q}_{m+l}\in\mathbb{R}^{d} \) denote the query vector embedded at position (m+l). Then attention score \({q}_{m+l}{k}_{m}\) is derived as follow

\begin{equation}
\begin{aligned}
{q}_{m+l} {k}_{m}
&={(\mathbb{R}^{d}_{\Theta, m+l}q)}^{T}{(\mathbb{R}^{d}_{\Theta, m}k)}\\
&= \sum_{i=0}^{d/2-1}({q_{[2i]}k_{[2i]}\cos (m+l-m)\theta_{i}}\\
& \quad +{q_{[2i+1]}k_{[2i+1]}\cos (m+l-m)\theta_{i}}) \\
&= \sum_{i=0}^{d/2-1}({q_{[2i]}k_{[2i]}+ {q_{[2i+1]}k_{[2i+1]}})\cos l\theta_{i}} \\
\end{aligned}
\end{equation}

where \(\{ q,k\}_{[i]}\) denotes i-th entry of vectors \(\{q,k\}\) and \( {h}_{i}\) denotes dot product \({q}_{i}{k}_{i}\). The attention score \({q}_{m+l}{k}_{m}\) only depends on the relative distance \(l\) rather than the absolute position \(m\).
\end{proposition}

\end{document}